\documentclass[letterpaper]{article} % DO NOT CHANGE THIS
\usepackage[preprint]{aaai2027}  % DO NOT CHANGE THIS
\usepackage[hyphens]{url}  % DO NOT CHANGE THIS
\usepackage{graphicx} % DO NOT CHANGE THIS
\usepackage{natbib}  % DO NOT CHANGE THIS AND DO NOT ADD ANY OPTIONS TO IT
\usepackage{caption} % DO NOT CHANGE THIS AND DO NOT ADD ANY OPTIONS TO IT
\usepackage{algorithm}
\usepackage{algorithmic}
\usepackage{amsmath,amssymb,amsfonts}
\usepackage{multirow}
\usepackage{makecell}
\usepackage{subcaption}
\usepackage{pdfpages}

\usepackage{newfloat}
\usepackage{listings}
\DeclareCaptionStyle{ruled}{labelfont=normalfont,labelsep=colon,strut=off} % DO NOT CHANGE THIS
\floatstyle{ruled}
\newfloat{listing}{tb}{lst}{}
\floatname{listing}{Listing}

\usepackage{booktabs}

\definecolor{promptcolor}{HTML}{BFC1C0}
\usepackage[most]{tcolorbox}
\tcbuselibrary{skins, breakable}
\newtcolorbox{promptbox}[1]{
    breakable,
    colback=white,
    colframe=promptcolor,
    arc=4pt,
    boxrule=0.8pt,
    title=#1,
    coltitle=white,
    colbacktitle=promptcolor,
    fonttitle=\bfseries
}

\title{Beyond Memory Majority: Latent-Source Reasoning for Multi-Agent Memory Arbitration}
\author{
    Chenchen Lin\textsuperscript{\rm 1}, Wenhao Yuan\textsuperscript{\rm 1}, Xuehe Wang\textsuperscript{\rm 2,3,4}, Edith Cheuk Han Ngai\textsuperscript{\rm 1}\corresponding 
}
\affiliations{
    \textsuperscript{\rm 1} Department of Electrical and Computer Engineering, The University of Hong Kong \\
    \textsuperscript{\rm 2} School of Artificial Intelligence, Sun Yat-sen University \\
    \textsuperscript{\rm 3} Southern Marine Science and Engineering Guangdong Laboratory (Zhuhai) \\
    \textsuperscript{\rm 4} Key Laboratory of Intelligent Assessment Technology for Sustainable Tourism, Ministry of Culture and Tourism, Sun Yat-sen University \\ 
}

\begin{document}

\maketitle

\begin{abstract}
Long-term multi-agent systems continuously accumulate the memories produced by different agents. Existing memory methods typically treat retrieved memories as independent evidence and combine them through voting or weighting. However, this independence assumption often fails in multi-agent settings: memories written by different agents may inherit the same upstream source or shared bias, causing correlated evidence to be repeatedly counted and creating a false majority. We term this failure mode \textit{Memory Correlation Bias}. To address the issue, we propose the \textbf{C}orrelation-\textbf{A}ware \textbf{M}emory \textbf{A}rbitration (CAMA) framework that jointly decouples retrieved memories and recovers missing independent evidence. We model the retrieved memories as query-conditioned evidence groups and combine neural dependency inference with provenance-based symbolic priors to estimate the effective number of independent evidence sources, thereby preventing correlated memories from forming a false majority. Since critical independent evidence may be absent from the initial retrieval set, \textsc{CAMA} further learns a sequential recovery policy that actively retrieves alternative evidence or traces upstream sources before making the final decision, aiming to recover sufficient independent evidence for reliable arbitration while minimizing retrieval cost. Experiments on multiple benchmarks demonstrate the superiority of our method over the state-of-the-art baseline methods, suppressing false majorities induced by correlated memories.
\end{abstract}


\section{Introduction}
Large Language Models (LLMs) are increasingly deployed as long-term multi-agent systems, where multiple agents collaborate over extended horizons and continuously write their observations, summaries, and reasoning results into shared persistent memory~\citep{rezazadeh2025collaborative, zhang2025g}. For query answering, such systems retrieve relevant memories and aggregate them into final decisions by treating retrieved entries as independent evidence and combining them through voting or weighting~\citep{yang2026auditing, ai2026beyond}. However, this independence assumption often fails in multi-agent settings: memories written by different agents may originate from the same upstream source or inherit shared biases, causing a single underlying evidential factor to be represented by multiple entries and repeatedly counted~\citep{kohli2026nine}. We term this failure mode \textit{Memory Correlation Bias}, where correlated memories inflate the perceived support for a hypothesis and lead to a \textit{false majority}. This problem is particularly harmful in long-term multi-agent systems: once a false majority determines the arbitration outcome, the erroneous conclusion is written back into shared memory as new evidence, further amplifying correlated signals and causing persistent, self-reinforcing errors in subsequent decisions~\citep{zhang2025g}.

In long-term agent systems, existing methods primarily aggregate retrieved memories through majority voting~\citep{yang2026auditing}, confidence or relevance weighting~\citep{ai2026beyond}, and retrieval-augmented reasoning~\citep{kang2025memory,yu2026agentic}. While improving robustness over single-memory reasoning, these approaches generally assume that retrieved memories provide independent evidence, such that more concordant entries indicate stronger support. In practice, however, whether two memories are redundant is \textit{query-specific}: memories from the same source may provide complementary evidence for one query while reinforcing the same factor for another, and such dependency cannot be captured solely by static attributes such as agent identity, semantic similarity, or provenance~\citep{liu2026consensus}. Recent studies have explored memory management~\citep{tan2025prospect}, provenance tracking, and reliability-aware aggregation~\citep{yang2026auditing}; however, query-conditioned redundancy among correlated memories and its impact on evidence aggregation remain largely unexplored. Further, the retrieved memory set is itself a biased subset of the memory store, as similarity-based retrieval tends to over-select correlated memories while under-selecting independent evidence that could resolve the false majority~\citep{salama2025meminsight}. Thus, critical independent evidence may be absent from the initial retrieval set, making it impossible for arbitration methods confined to the retrieved memories to recover the correct decision. These observations reveal that reliable memory arbitration should account for the effective independence of evidence sources rather than their raw frequency, raising a key question: \textit{How can a long-term multi-agent system arbitrate conflicting memories by their effectively independent evidence rather than their apparent count?}

To overcome these limitations, we propose \textbf{C}orrelation-\textbf{A}ware \textbf{M}emory \textbf{A}rbitration (\textsc{CAMA}), a framework that models query-conditioned evidential dependencies and recovers missing independent evidence before making a decision. To model correlated evidence, we represent retrieved memories as query-conditioned latent evidence slots and combine neural dependency inference with provenance-based symbolic priors to estimate the effective number of independent evidence sources, preventing correlated memories from repeatedly counting the same factor toward a false majority. Then, \textsc{CAMA} aggregates hypothesis support at the level of latent evidence factors rather than individual entries, attributing decisions to reliable and effectively independent sources. Given that critical independent evidence may be absent from the initial retrieval, we learn a sequential recovery policy that actively expands the retrieval space or traces memory dependencies to acquire additional independent evidence while minimizing recovery cost. Our key contributions are summarized as follows:
\begin{itemize}
\item We identify the overlooked problem of \textit{Memory Correlation Bias} in long-term multi-agent systems, where correlated memories are repeatedly counted as independent evidence and form a false majority.

\item We introduce \textsc{CAMA}, a novel framework that decouples correlated memories into effectively independent evidence sources, arbitrates conflicts at the evidence-factor level, and actively recovers missing independent evidence under a retrieval budget.

\item We conduct extensive experiments on multiple benchmarks, demonstrating that \textsc{CAMA} outperforms state-of-the-art baselines and effectively suppresses false majorities induced by correlated memories.
\end{itemize}

\section{Related Work}

\subsection{Memory in Long-Term Multi-Agent Systems}
Persistent memory has become a central component for adapting LLM-based multi-agent systems to long-horizon collaboration~\citep{huang2026ama}. Existing work studies how agents write, organize, and retrieve shared memories over extended interactions, showing that persistent memory improves continuity, coordination, and downstream task performance~\citep{huang2026ama, zhang2025survey}. Subsequent methods improve memory utility through memory management and updating~\citep{yu2026agentic,xu2025mem}, retrieval-augmented memory reasoning~\citep{xu2026chain,du2025memr}, and provenance tracking or auditing of stored evidence~\citep{souza2025prov,wang2026agent}. To reach a final decision, such systems typically aggregate retrieved memories, combining concordant entries or agent outputs through voting and confidence- or relevance-based weighting~\citep{lu2026mma, liang2025cognitive}. These works demonstrate the importance of accumulating and exploiting historical memory, especially when relevant evidence is distributed across many agents and interactions.

Despite the advances, most methods treat retrieved memories as independent evidence, equating greater agreement with stronger support. In multi-agent settings, redundancy is \textit{query-specific}: memories from the same source may be complementary for one query but reinforce the same factor for another, a dependency that static attributes such as agent identity, semantic similarity, or provenance cannot capture~\citep{liu2026consensus, kohli2026nine}. Thus, correlated memories may be repeatedly counted, inflating support for a hypothesis.

\subsection{Evidence Aggregation and Recovery}
A parallel line of work seeks to improve evidence aggregation beyond naive counting. To reduce unreliable signals, existing methods use consistency-based aggregation over multiple candidates~\citep{taubenfeld2025confidence}, source reliability estimation, and confidence-aware weighting~\citep{hwang2025retrieval, razghandi2025cer}. Other studies examine correlations and conflicts across evidence sources~\citep{kim2025correlated, ge2025resolving}, while neuro-symbolic methods incorporate structural priors into evidence reasoning~\citep{peer2025ata,yang2025neuro}. Redundancy is also mitigated through semantic or provenance-based filtering~\citep{chang2025main,peng2025cafe}. When the initial evidence is insufficient, iterative and retrieval-augmented reasoning methods recover additional evidence through further queries~\citep{lin2025rje,tran2025rare}. Together, these methods improve aggregation by modeling reliability, dependency, redundancy, and evidence coverage.

Despite this progress, existing aggregation methods either assume independent evidence or rely on static dependency structures, failing to capture query-conditioned memory correlations. Redundancy reduction mainly relies on similarity or provenance rather than the effective number of independent sources, while recovery methods may introduce correlated memories without resolving dependencies. Thus, they cannot jointly decouple correlations and recover missing independent evidence, leaving false majorities unresolved.

\begin{figure*}[t]
\centering
\includegraphics[width=1.0\textwidth]{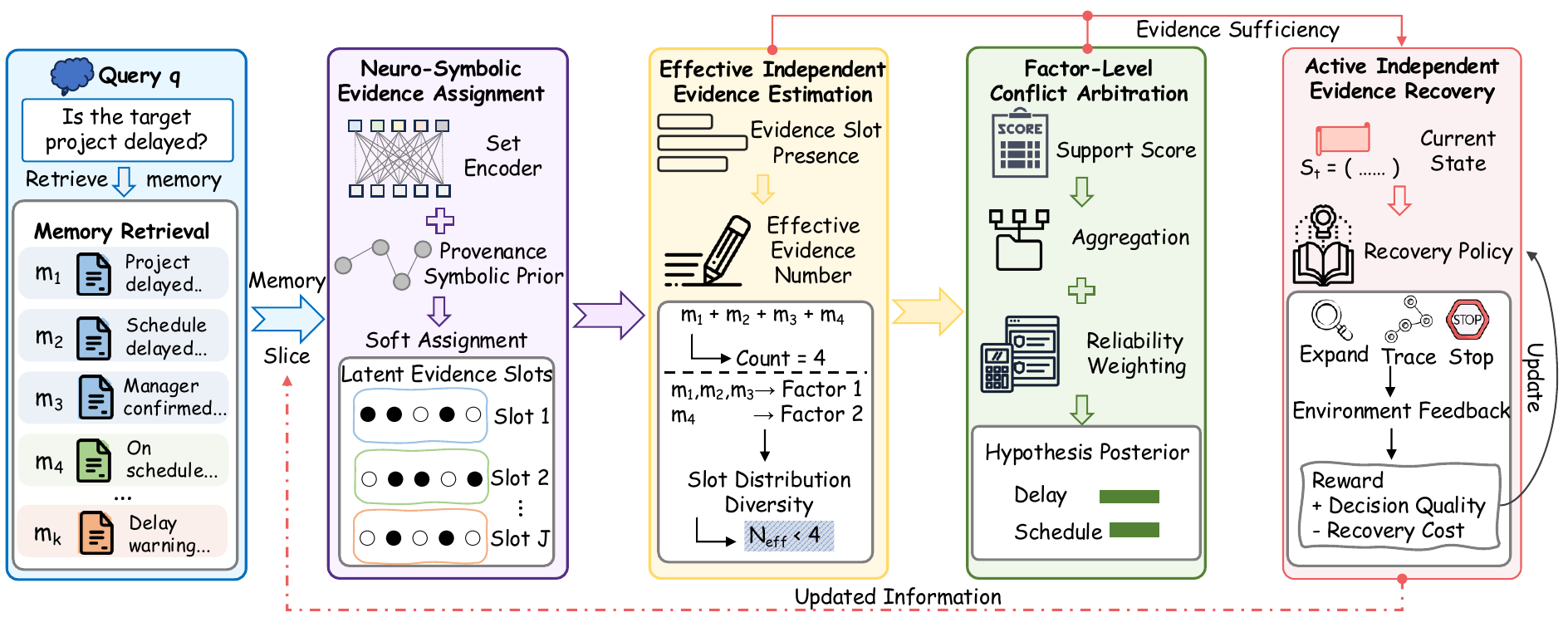}
\caption{An overview of our proposed \textsc{CAMA}. The diagram illustrates the overall workflow of memory arbitration, where retrieved memories are progressively processed through evidence decoupling, conflict arbitration, and evidence recovery.}
\label{framework}
% \vspace{-10pt}
\end{figure*}

% We propose Correlation-Aware Memory Arbitration (CAMA), a framework for resolving conflicting memories in long-term multi-agent systems. Given a retrieved memory slice, \textsc{CAMA} infers soft assignments between memories and latent, query-conditioned evidence clusters, enabling arbitration based on effectively independent evidence sources rather than individual memory entries. When the retrieved evidence is insufficient, \textsc{CAMA} further performs active evidence recovery by expanding the retrieval space or tracing memory dependencies to acquire additional independent support. The recovered evidence is then incorporated into the arbitration process, allowing the model to update its dependency estimation and progressively refine the final decision.

\section{Methodology} \label{sec:methodology}

\subsection{Problem Formulation} \label{sec:problem-formulation}

As illustrated in Figure~\ref{framework}, we consider a multi-agent system $\mathcal{A}=\{A_1,\ldots,A_N\}$ with a memory store $\mathcal{M}= \bigcup_{j=1}^{N}\mathcal{M}_j$, where $\mathcal{M}_j$ contains the observations, summaries, intermediate reasoning results, and execution traces generated by $A_j$. Given a query $q$, a retrieval module returns an initial memory slice $\mathcal{C}^{(0)}_q=\operatorname{Retrieve}(q,\mathcal{M};K)$, where $|\mathcal{M}|\gg K$. During evidence recovery, \textsc{CAMA} iteratively updates the memory slice and denotes the state after $t$ recovery steps as $\mathcal{C}^{(t)}_q$. Based on the current slice, we maintain a candidate hypothesis set $\mathcal{H}^{(t)}_q=\{h^{(t)}_1,\ldots,h^{(t)}_{L_t}\}$, where newly recovered memories may introduce additional hypotheses. The candidate extraction process is orthogonal to \textsc{CAMA}, which focuses on evidence modeling and memory arbitration.

Unlike conventional aggregation methods that treat retrieved memories as independent evidence, \textsc{CAMA} models memory dependence through \textit{query-conditioned evidential redundancy}. Specifically, two memories $m_i$ and $m_j$ are considered dependent under query $q$ if they share the same latent evidential factor or provide overlapping support for the same underlying evidence, denoted as $\operatorname{Dep}(m_i,m_j|q)$. Such dependence is query-specific and cannot be fully characterized by static attributes such as agent identity, semantic similarity, or provenance, since different memories may either provide distinct evidence from the same source or propagate the same evidence across different agents. Therefore, \textsc{CAMA} focuses on inferring the effective evidential relationships among memories relevant to the current decision rather than maintaining a global dependency structure.

Given a recovery budget $B$, \textsc{CAMA} determines the final hypothesis $\hat{h}$ through a sequence of evolving memory states $\{\mathcal{C}^{(t)}_q\}_{t=0}^{T}$, where $T\leq B$. The objective is to perform evidence-aware arbitration by avoiding redundant counting of correlated memories and recovering missing independent evidence when the initial retrieval is insufficient.

\subsection{Query-Conditioned Evidence Decoupling} \label{sec:evidence-decoupling}
Evidence decoupling aims to identify effectively independent evidence sources underlying the current memory slice rather than counting retrieved entries. \textsc{CAMA} models these sources as latent query-conditioned evidence slots and infers soft memory-to-slot assignments via a neuro-symbolic module that integrates provenance-based priors with a set encoder. Provenance serves as supporting evidence rather than a hard dependency label, enabling query-dependent redundancy modeling \cite{garcez2015neural,hitzler2022neural}.

\subsubsection{Neuro-Symbolic Evidence Assignment}
Given the current memory slice $\mathcal{C}^{(t)}_q$, the assignment module infers soft assignments between memories and latent evidence slots. Let $J$ denote the maximum number of latent evidence slots, with inactive slots automatically ignored, and let $G^{(t)}_{\mathrm{prov}}$ denote the temporary provenance graph constructed from the current slice when provenance metadata is available. For $K_t=|\mathcal{C}^{(t)}_q|$ memories, a set-based self-attention encoder jointly models the query, memory interactions, and provenance structure to produce the assignment matrix
\begin{align}
Z^{(t)}=f_{\Theta}(q,\mathcal{C}^{(t)}_q,G^{(t)}_{\mathrm{prov}})\in\Delta^{K_t\times J},
\end{align}
where $\Delta^{K_t\times J}$ denotes the space of row-wise probability distributions, and each row $\mathbf{z}^{(t)}_i=(z^{(t)}_{i1},\ldots,z^{(t)}_{iJ})$ represents the soft assignment distribution of memory $m_i$ over the latent evidence slots. Unlike one-to-one memory clustering, the soft assignment allows each memory to reflect multiple evidential factors and captures partial dependence among memories. Since redundancy depends on both the query and the memory context, the assignment of each memory is inferred jointly from the entire slice rather than independently from individual memory content, further providing a query-conditioned overlap measure $r^{(t)}_{ij}=\langle\mathbf{z}^{(t)}_i,\mathbf{z}^{(t)}_j\rangle$, where a larger value indicates that two memories share similar latent evidence factors under the current query. Provenance information is incorporated as a symbolic prior to guide evidence assignment rather than enforcing hard dependency constraints. Specifically, provenance relations in $G^{(t)}_{\mathrm{prov}}$ are injected into the self-attention mechanism as biases:
\begin{align}
a_{ij}=\frac{\mathbf{q}_i^{\top}\mathbf{k}_j}{\sqrt{d}}+\mu b_{ij},
\end{align}
where $\mathbf{q}_i,\mathbf{k}_j\in\mathbb{R}^{d}$ denote the query and key representations of memories $m_i$ and $m_j$, $b_{ij}$ encodes the observed provenance relation, and $\mu $ controls the strength of the symbolic prior. This prior encourages interactions among structurally related memories, while the final assignments remain determined by the query-conditioned neural representations. The confidence of the inferred assignments is quantified by the normalized entropy of each slot distribution:
\begin{align}
\kappa^{(t)}_i = 1 + \frac{1}{\log J} \sum_{j=1}^{J}z^{(t)}_{ij}\log z^{(t)}_{ij},
\end{align}
measuring confidence of the inferred evidence assignment, with larger values indicating lower assignment uncertainty.

\subsubsection{Effective Independent Evidence Estimation}
The soft assignments characterize how retrieved memories contribute to latent evidence slots. However, the number of memory entries does not necessarily reflect the amount of independent evidence, as multiple memories may originate from the same underlying evidential factor. To estimate the presence of each evidence slot, we use the strongest assignment among the retrieved memories $e^{(t)}_j=\max_{1\leq i\leq K_t}z^{(t)}_{ij}$, where $e^{(t)}_j\in[0,1]$ measures the extent to which evidence slot $j$ is represented in $\mathcal{C}^{(t)}_q$. This max-based definition prevents correlated memories from increasing evidence mass through repeated representations of the same factor. Based on the slot-presence values, the effective number of independent evidence sources is quantified using a Hill diversity measure. Specifically, the slot presence is first normalized as $p^{(t)}_j= \frac{e^{(t)}_j}{\sum_{l=1}^{J}e^{(t)}_l}$, where $p^{(t)}_j$ denotes the relative presence of evidence slot $j$. The effective evidence number is then defined as
% \begin{align} \label{eq:effective-evidence-number}
% N^{(t)}_{\mathrm{eff}} = \left(\sum_{j=1}^{J}(p^{(t)}_j)^{\alpha}
% \right)^{\frac{1}{1-\alpha}},
% \end{align}
\begin{align}
N^{(t)}_{\mathrm{eff}} =
\exp\left(
\frac{\log\sum_{j=1}^{J}(p_j^{(t)})^\alpha}
{1-\alpha}
\right).
\end{align}
where $\alpha$ denotes the diversity order. A larger $N^{(t)}_{\mathrm{eff}}$ indicates that the retrieved memories cover a more diverse set of latent evidence factors, whereas a smaller value reflects evidence concentration on fewer factors. By operating on slot-level presence rather than memory frequency, this measure captures effective independent evidence and is used for subsequent arbitration and evidence recovery.

\subsection{Factor-Level Conflict Arbitration} \label{sec:cluster-arbitration}
The inferred latent evidence structure provides a basis for hypothesis-level arbitration, where competing hypotheses are evaluated based on evidence factors rather than individual memory entries. For each memory $m_i$ and candidate hypothesis $h\in\mathcal{H}^{(t)}_q$, a candidate-conditioned scorer produces a support score $s^{(t)}_i(h)=s_{\Theta}(\mathbf{u}^{(t)}_i,h)$, where $\mathbf{u}^{(t)}_i$ denotes the query- and set-conditioned representation of memory $m_i$. By taking the candidate hypothesis as an input, the same scorer can evaluate newly introduced hypotheses during subsequent evidence recovery steps without modifying the output space. The support associated with evidence slot $j$ is then aggregated over the memories assigned to that slot:
\begin{align} \label{eq:cluster-support}
\beta^{(t)}_j(h) = \frac{\sum_{i=1}^{K_t}z^{(t)}_{ij}s^{(t)}_i(h)}{\sum_{i=1}^{K_t}z^{(t)}_{ij}+\epsilon},
\end{align}
where the normalization ensures that an evidence source does not gain additional influence simply because it has generated more memory descendants. $\beta^{(t)}_j(h)$ measures the support of evidence slot $j$ for hypothesis $h$. To further account for variations in source reliability, memory-level prior attributes are aggregated within each evidence slot to estimate a reliability weight. For each memory $m_i$, we define $\mathbf{a}_{\mathrm{prior}}(m_i) = \left[ o_i, r_i, \mathbf{\chi}_i \right] \in\mathbb{R}^{d_a}$, where $o_i\in[0,1]$ indicates whether $m_i$ originates from a direct observation, $r_i\in[0,1]$ denotes the historical reliability of its generating agent, and $\mathbf{\chi}_i$ is a one-hot encoding of the upstream source type. The slot-level reliability weights are computed as
\begin{align} \label{eq:cluster-reliability}
\rho^{(t)}_j =\sigma\left( \mathbf{w}_{\rho}^{\top} (\frac{\sum_i z^{(t)}_{ij}\mathbf{a}_{\mathrm{prior}}(m_i)}{\sum_i z^{(t)}_{ij}+\epsilon}) + b_{\rho} \right),
\end{align}
where $\mathbf{w}_{\rho}$ and $b_{\rho}$ are learnable parameters, and $\sigma(\cdot)$ denotes the sigmoid function. The normalized aggregation prevents the estimated reliability from being biased by memory multiplicity. When source metadata is unavailable, we use a learnable default reliability weight $\rho^{(t)}_j=\sigma(b_{\mathrm{miss}})$. The evidence contribution of each latent factor is then aggregated into the arbitration logit $\ell^{(t)}(h)=\sum_{j=1}^{J}\rho^{(t)}_j e^{(t)}_j\beta^{(t)}_j(h)$, where $\rho^{(t)}_j$ and $e^{(t)}_j$ regulate the contribution of each evidence factor based on its trustworthiness and availability. The arbitration posterior is obtained by temperature-scaled normalization:
\begin{align} \label{eq:arbitration-posterior}
P^{(t)}(h\mid q,\mathcal{C}^{(t)}_q) = \frac{\exp(\ell^{(t)}(h)/\tau_p)}{\sum_{h'\in\mathcal{H}^{(t)}_q}\exp(\ell^{(t)}(h')/\tau_p)},
\end{align}
where $\tau_p>0$ is calibrated on validation data. Reusing $e^{(t)}_j$ in both evidence estimation and arbitration ensures that each evidential factor contributes according to its presence and reliability rather than memory frequency. The evidence sufficiency is jointly assessed by $N^{(t)}_{\mathrm{eff}}$ and the arbitration posterior, which capture evidence diversity and hypothesis separation, to determine whether further recovery is required.

\subsection{Active Independent-Evidence Recovery} \label{sec:active-recovery}
The arbitration process relies on the evidence available in the current memory slice. When the retrieved memories provide insufficient independent evidence or contain unresolved dependencies, additional evidence recovery is required. We formulate recovery as a finite-horizon sequential decision process over the memory store, where the policy selects among evidence expansion, dependency tracing, and termination actions. At recovery step $t$, the state is defined as
\begin{align}
S_t=(\mathcal{C}^{(t)}_q,G^{(t)}_{\mathrm{prov}},Z^{(t)},N^{(t)}_{\mathrm{eff}},\boldsymbol{\kappa}^{(t)},P^{(t)},t),
\end{align}
which summarizes the current memory slice, inferred evidence structure, evidence sufficiency, assignment confidence, and arbitration uncertainty. The policy selects an action from $A_t\in\left\{\textsc{Expand}(q'),\textsc{Trace}(m_i),\textsc{Stop}\right\}$:
\begin{itemize}
\item $\textsc{Expand}(q')$: This action recovers independent evidence that may be missing from the current retrieval view. Specifically, the policy generates a bounded set of alternative query reformulations from the current state and selects one to retrieve additional memories, updated as $\mathcal{C}^{(t+1)}_q=\mathcal{C}^{(t)}_q\cup\operatorname{Retrieve}(q',\mathcal{M};K_{\mathrm{add}})$.

\item $\textsc{Trace}(m_i)$: This action follows a recorded derivation edge from memory $m_i$ to its parent memory $m_p$. The recovered parent and provenance relation are added to the current slice and local provenance graph, i.e., $\mathcal{C}^{(t+1)}_q=\mathcal{C}^{(t)}_q\cup\{m_p\}$ and $G^{(t+1)}_{\mathrm{prov}}=G^{(t)}_{\mathrm{prov}}\cup\{m_i\leftarrow m_p\}$. If memories share the recovered parent, their corresponding edges are added simultaneously. The set-conditioned assignments are then recomputed over the updated slice and graph. \textsc{Trace} can reveal that memories previously treated as independent originate from the same upstream evidence, reducing $N^{(t)}_{\mathrm{eff}}$ and improving arbitration.

\item $\textsc{Stop}$: This action terminates recovery and returns the current arbitration decision $\hat h = \arg\max_{h\in\mathcal{H}^{(t)}_q}P^{(t)}(h\mid q,\mathcal{C}^{(t)}_q)$. The termination decision is guided by both the learned policy and an interpretable sufficiency criterion $N^{(t)}_{\mathrm{eff}}\geq\tau_N$ and $H(P^{(t)})\leq\tau_H$. The first condition requires sufficient independent evidence, while the second requires a concentrated arbitration posterior. The thresholds are calibrated on validation data, and all policies terminate when the recovery budget $B$ is exhausted.
\end{itemize}

When provenance information is unavailable, the recovery process uses only \textsc{Expand} and \textsc{Stop}. Query reformulation can still retrieve complementary evidence, while \textsc{Trace} requires explicit provenance signals.

\subsection{Evidence-Guided Recovery Optimization} \label{sec:policy-learning}
Given the state representation, the policy $\pi_{\omega}(A_t\mid S_t)$ is optimized to balance evidence acquisition and recovery cost under a budget. The policy determines whether to acquire additional evidence, investigate potential dependencies, or terminate with the current arbitration result. We optimize the policy with a terminal-oriented reward: nonterminal recovery actions incur only memory-access costs, i.e., $R_t=-\lambda_s$ for $A_t\in\{\textsc{Expand},\textsc{Trace}\}$, while stopping receives:
\begin{align}
R_t=
\begin{cases}
  +1, & \hat h=h^*,\\
  -1, & \hat h\neq h^*,
\end{cases}
\quad A_t=\textsc{Stop},
\label{eq:terminal-reward}
\end{align}
where $h^*$ denotes the ground-truth conclusion during training. The terminal-oriented reward evaluates the final arbitration outcome while accounting for memory access costs during recovery. Since the effect of a recovery action may emerge after subsequent evidence updates, its utility is learned through long-term returns. To make value estimation evidence-aware, the current state is summarized using correlation-aware statistics:
\begin{align}
    \mathbf{d}_t = \left[
      N^{(t)}_{\mathrm{eff}},
      \overline\kappa^{(t)},
      H(P^{(t)}),
      p^{(t)}_{(1)},
      p^{(t)}_{(1)}-p^{(t)}_{(2)},
      t/B
    \right],
    \label{eq:critic-summary}
\end{align}
where $p_{(1)}$ and $p_{(2)}$ denote the two largest hypothesis probabilities, and $\overline\kappa^{(t)}=\frac{1}{K_t}\sum_{i=1}^{K_t}\kappa^{(t)}_i$. The value function $V_{\nu}(\mathbf{d}_t)$ estimates the expected arbitration quality from effective evidence quantity, assignment confidence, and posterior uncertainty. The actor uses the contextual state representation to select recovery actions. Given return $G_t=\sum_{k\geq 0}\gamma^kR_{t+k}$ and advantage estimate $\widehat A_t=G_t-V_{\nu}(\mathbf{d}_t)$, the policy and value function are optimized with an actor--critic objective:
\begin{align}
    \mathcal{L}_{\mathrm{A}}
    &= -\mathbb{E}_t\left[
       \log\pi_{\omega}(A_t\mid S_t)
       \operatorname{sg}(\widehat A_t)\right],
    \label{eq:actor-loss}\\
    \mathcal{L}_{\mathrm{V}}
    &= \mathbb{E}_t\left[
       \left(V_{\nu}(\mathbf{d}_t)-\operatorname{sg}(G_t)\right)^2
       \right],
    \label{eq:value-loss}\\
    \mathcal{L}_{\mathrm{RL}}
    &= \mathcal{L}_{\mathrm{A}}
       +c_v\mathcal{L}_{\mathrm{V}}
       -c_e\,\mathbb{E}_t[H(\pi_{\omega}(\cdot\mid S_t))],
    \label{eq:rl-loss}
\end{align}
where $\operatorname{sg}(\cdot)$ denotes stop-gradient. The evidence representation underlying state construction and arbitration is jointly optimized with the recovery policy to support reliable decisions. Specifically, the shared encoder and prediction heads are trained with the arbitration objective and, when provenance information is available, an auxiliary dependence-aware contrastive objective. For the final slice of a training episode, the arbitration loss is defined as
\begin{align}
    \mathcal{L}_{\mathrm{task}}
    = -\log P^{(T)}(h^*\mid q,\mathcal{C}^{(T)}_q),
    \label{eq:task-loss}
\end{align}
where $h^*$ denotes the ground-truth conclusion. To leverage provenance information, a weakly supervised contrastive objective is introduced over evidence assignments. For each anchor memory $i$, $\mathcal{P}(i)$ and $\mathcal{N}(i)$ denote memories with observed provenance relations and distinct source origins, respectively. Since provenance provides partial dependency cues, it is used as weak supervision rather than a hard assignment constraint. The dependence loss is defined as:
\begin{align} \label{eq:dependence-loss}
\!\! \mathcal{L}_{\mathrm{d}} \!=\! -\mathbb{E}_{i\sim\mathcal B}
\mathbb{E}_{j\sim\mathcal P(i)}
\log
\frac{\exp(r^{(T)}_{ij}/\tau_d)}
{\sum_{k\in\mathcal P(i)\cup\mathcal N(i)}
\exp(r^{(T)}_{ik}/\tau_d)} . \!\!
\end{align}
where $r^{(T)}_{ij}=\langle \mathbf{z}^{(T)}_i,\mathbf{z}^{(T)}_j\rangle$ measures the overlap between the source assignment distributions of memories $i$ and $j$, and $\tau_d$ is the temperature parameter. Provenance-based contrastive supervision regularizes evidence assignments without relying on semantic similarity as a proxy for dependency. The overall training objective is
\begin{align}
\mathcal{L} = \mathcal{L}_{\mathrm{RL}} +\lambda_{\mathrm{task}}\mathcal{L}_{\mathrm{task}} +\lambda_{\mathrm{d}}\mathcal{L}_{\mathrm{d}}.
\end{align}

% As defined in Eqs.~\eqref{eq:actor-loss}--\eqref{eq:value-loss}, returns and advantage estimates are detached during policy optimization. The evidence encoder is updated through the differentiable state representations provided to the actor and critic, including $N_{\mathrm{eff}}$, $\overline\kappa$, and $P$, as well as through the supervised arbitration objectives in Eq.~\eqref{eq:task-loss} and Eq.~\eqref{eq:dependence-loss}.

\begin{table*}[t]
\centering
\renewcommand\arraystretch{0.9}
\resizebox{1.0\textwidth}{!}{
\begin{tabular}
{c|c|ccc|cccc|cccc} 
\toprule[1.2pt]
\multirow{2}{*}{\multirowcell{1.5}{\centering\textbf{Backbone}}} & \multirow{2}{*}{\multirowcell{1.5}{\centering\textbf{Methods}}} & \multicolumn{3}{c|}{\centering\textbf{MemoryAgentBench}} & \multicolumn{4}{c|}{\centering\textbf{LongMemEval}} & \multicolumn{4}{c}{\centering\textbf{LOCOMO}}  \\ \cmidrule[0.5pt](l{1pt}r{0pt}){3-13}

& & FC-SH & FC-MH & Overall & EM & F1 & BERT & Judge & EM & F1 & BERT & Judge  \\ \cmidrule[0.8pt](l{1pt}r{0pt}){1-13}

\multirow{6}{*}{\multirowcell{2}{DeepSeek-V4-Flash}}
        & \textsc{Vanilla RAG} & 68.4 & 34.2 & 51.3 & 34.1 & 45.7 & 84.2 & 51.6 & 29.7 & 40.3 & 83.5 & 47.2 \\
        
        & \textsc{Majority Voting} & 70.1 & 33.5 & 51.8 & 33.6 & 45.1 & 84.0 & 50.9 & 28.9 & 39.6 & 83.2 & 46.1 \\
        
        & \textsc{HippoRAG} & 72.6 & 42.8 & 57.7 & 38.5 & 49.8 & 85.3 & 56.4 & 33.4 & 44.1 & 84.6 & 51.8 \\
        
        & \textsc{Mem0} & 73.9 & 44.5 & 59.2 & 40.2 & 51.6 & 85.7 & 58.1 & 35.8 & 46.7 & 85.2 & 54.3 \\

        & \textsc{MAD} & 74.7 & 46.9 & 60.8 & 41.7 & 52.9 & 86.1 & 60.5 & 36.9 & 47.8 & 85.5 & 56.2 \\
        
        & \textsc{MADAM-RAG} & 75.2 & 48.8 & 63.4 & 44.1 & 54.3 & 86.8 & 62.4 & 37.6 & 49.5 & 85.7 & 59.4 \\
        
        & \textsc{CAMA} (Ours) & \textbf{78.9} & \textbf{55.7} & \textbf{67.3} & \textbf{49.8} & \textbf{59.1} & \textbf{87.9} & \textbf{69.2} & \textbf{43.6} & \textbf{53.8} & \textbf{87.1} & \textbf{64.7} \\ \midrule[0.8pt]

\multirow{6}{*}{\multirowcell{2}{Qwen3.6-27B}}
        & \textsc{Vanilla RAG} & 65.2 & 31.4 & 48.3 & 31.5 & 42.9 & 83.4 & 48.7 & 27.3 & 37.8 & 82.7 & 44.5 \\
        
        & \textsc{Majority Voting} & 66.8 & 30.7 & 48.8 & 30.9 & 42.3 & 83.1 & 47.9 & 26.5 & 37.1 & 82.4 & 43.6 \\
        
        & \textsc{HippoRAG} & 69.5 & 39.6 & 54.6 & 35.6 & 46.8 & 84.5 & 53.2 & 30.9 & 41.4 & 83.8 & 49.1 \\
        
        & \textsc{Mem0} & 71.0 & 41.3 & 56.2 & 37.4 & 48.7 & 84.9 & 55.3 & 33.2 & 43.9 & 84.4 & 51.7 \\

        & \textsc{MAD} & 71.9 & 43.8 & 57.9 & 38.9 & 50.1 & 85.3 & 57.6 & 34.5 & 45.2 & 84.7 & 53.8 \\
        
        & \textsc{MADAM-RAG} & 73.6 & 46.7 & 60.2 & 41.6 & 52.6 & 85.9 & 61.2 & 36.8 & 47.5 & 85.3 & 57.1 \\
        
        & \textsc{CAMA} (Ours) & \textbf{76.5} & \textbf{53.2} & \textbf{64.9} & \textbf{47.4} & \textbf{56.9} & \textbf{87.2} & \textbf{67.0} & \textbf{41.5} & \textbf{51.6} & \textbf{86.5} & \textbf{62.4} \\

\bottomrule[1.2pt]
\end{tabular}}
\caption{Overall performance comparison on three benchmark datasets. Best results are marked by \textbf{bold}.}
\label{accuracy}
% \vspace{-4pt}
\end{table*}

\begin{table*}[t]
\centering
\renewcommand\arraystretch{0.9}
\resizebox{1.0\textwidth}{!}{
\begin{tabular}{c|cccc|cccc|cccc} 
\toprule[1.2pt]
\multirow{2}{*}{\multirowcell{2}{\centering\textbf{Methods}}} & \multicolumn{4}{c|}{\centering\textbf{MemoryAgentBench}} & \multicolumn{4}{c|}{\centering\textbf{LongMemEval}} & \multicolumn{4}{c}{\centering\textbf{LOCOMO}}  \\ \cmidrule[0.5pt](l{1pt}r{0pt}){2-13}

& CMR $\uparrow$ & RS $\downarrow$ & IEG $\uparrow$ & ERR $\uparrow$ & CMR $\uparrow$ & RS $\downarrow$ & IEG $\uparrow$ & ERR $\uparrow$ & CMR $\uparrow$ & RS $\downarrow$ & IEG $\uparrow$ & ERR $\uparrow$ \\ \cmidrule[0.8pt](l{1pt}r{0pt}){1-13}

\textsc{Vanilla RAG} & 38.7 & 41.2 & 5.8 & 5.3 & 36.9 & 43.5 & 5.1 & 4.6 & 33.4 & 45.8 & 4.4 & 3.9 \\

\textsc{Majority Voting} & 33.5 & 44.8 & 4.7 & 4.9 & 31.8 & 46.9 & 4.2 & 4.2 & 28.7 & 49.3 & 3.5 & 3.5 \\

\textsc{HippoRAG} & 46.8 & 27.4 & 11.6 & 9.8 & 44.2 & 29.1 & 10.5 & 8.9 & 40.5 & 31.6 & 9.1 & 7.7 \\

\textsc{Mem0} & 49.6 & 24.1 & 13.7 & 11.2 & 47.1 & 25.8 & 12.6 & 10.3 & 43.4 & 28.2 & 11.0 & 9.0 \\

\textsc{MAD} & 54.3 & 19.2 & 15.9 & 12.6 & 51.8 & 20.7 & 14.5 & 11.4 & 47.6 & 22.9 & 12.8 & 10.1 \\

\textsc{MADAM-RAG} & 60.7 & 15.3 & 19.4 & 14.1 & 58.2 & 16.6 & 18.1 & 12.9 & 53.9 & 18.5 & 16.2 & 11.5 \\ \midrule[0.8pt]

\textsc{CAMA} (Ours) & \textbf{71.2} & \textbf{7.8} & \textbf{25.1} & \textbf{36.2} & \textbf{67.4} & \textbf{9.1} & \textbf{22.6} & \textbf{33.1} & \textbf{62.1} & \textbf{10.3} & \textbf{20.2} & \textbf{29.4} \\

\bottomrule[1.2pt]
\end{tabular}}
% }
\caption{Evaluation of memory correlation bias mitigation under DeepSeek-V4-Flash.}
\label{correlation_bias}
% \vspace{-10pt}
\end{table*}

\section{Experiments}

\subsection{Experimental Setup}

\paragraph{Datasets}
We evaluate \textsc{CAMA} on three long-term memory benchmarks:
\textbf{MemoryAgentBench}~\citep{hu2026evaluating} evaluates LLM agent memory capabilities through incremental multi-turn interactions, including selective forgetting under conflicting memory updates. \textbf{LongMemEval}~\citep{wu2025longmemeval} evaluates long-term memory retrieval and reasoning in conversational agents under extended interaction histories. \textbf{LoCoMo}~\citep{maharana2024evaluating} focuses on long-term conversational memory reasoning over multi-session dialogues with evolving user states and historical interactions. To evaluate memory correlation bias, we construct correlation-aware variants from the original instances by augmenting memory pools with correlated entries from shared evidence sources and complementary entries providing distinct query-relevant evidence. Correlated entries are generated through controlled derivations (e.g., paraphrasing and summarization), with provenance links recorded while preserving the original ground-truth answers.

\paragraph{Baselines}
We compare \textsc{CAMA} with state-of-the-art baselines. \textbf{Vanilla RAG}~\citep{lewis2020retrieval} directly conditions generation on retrieved memories while treating all retrieved entries as independent evidence. \textbf{Majority Voting}~\citep{wang2023selfconsistency} aggregates memories based on consensus, representing conventional evidence aggregation strategies that assume each memory contributes an independent vote. We further evaluate long-term memory systems, including \textbf{Mem0}~\citep{chhikara2025mem0}, which extracts, consolidates, and updates salient memories for scalable memory management, and \textbf{HippoRAG}~\citep{gutierrez2024hipporag}, which employs graph-based memory organization to support long-range retrieval. For multi-agent scenarios, we compare with \textbf{MAD}~\citep{liang2024encouraging}, which improves reasoning through iterative interactions among multiple agents, and \textbf{MADAM-RAG}~\citep{wang2025retrieval}, which addresses conflicting evidence through multi-agent retrieval and aggregation.

\begin{table*}[t]
\centering
\renewcommand\arraystretch{0.9}
\resizebox{1.0\textwidth}{!}{
\begin{tabular}{c|ccccccc|cccccccc} 
\toprule[1.2pt]
\multirow{2}{*}{\multirowcell{2}{\centering\textbf{Methods}}} & \multicolumn{7}{c|}{\centering\textbf{MemoryAgentBench}} & \multicolumn{8}{c}{\centering\textbf{LongMemEval}} \\ \cmidrule[0.5pt](l{1pt}r{0pt}){2-16}

& FC-SH & FC-MH & Overall & CMR & RS & IEG & ERR & EM & F1 & BERT & Judge & CMR & RS & IEG & ERR  \\ \cmidrule[0.8pt](l{1pt}r{0pt}){1-16}

w/o Evi. Decoupling & 71.5 & 45.3 & 58.4 & 48.2 & 32.7 & 14.6 & 28.9 & 42.1 & 51.8 & 86.2 & 59.7 & 46.5 & 34.1 & 13.8 & 26.4 \\ % \cmidrule[0.5pt](l{1pt}r{0pt}){2-13}

w/o Prov. Prior & 76.8 & 51.9 & 64.4 & 63.4 & 13.8 & 21.7 & 34.2 & 47.3 & 56.6 & 87.4 & 65.8 & 61.2 & 14.6 & 20.3 & 31.8 \\

w/o Expand & 76.2 & 51.4 & 63.8 & 66.7 & 9.5 & 17.2 & 15.3 & 46.1 & 55.7 & 87.3 & 64.9 & 63.8 & 10.2 & 15.8 & 13.6 \\

w/o Trace & 75.4 & 49.6 & 62.5 & 64.8 & 15.2 & 19.4 & 27.1 & 46.8 & 56.2 & 87.4 & 65.7 & 62.5 & 16.3 & 18.1 & 25.2 \\

w/o Policy & 77.1 & 52.3 & 64.7 & 65.9 & 11.4 & 16.8 & 22.7 & 47.6 & 57.0 & 87.5 & 66.4 & 64.1 & 12.5 & 15.2 & 20.8 \\
\midrule[0.8pt]

\textsc{CAMA} & \textbf{78.9} & \textbf{55.7} & \textbf{67.3} & \textbf{71.2} & \textbf{7.8} & \textbf{25.1} & \textbf{36.2} & \textbf{49.8} & \textbf{59.1} & \textbf{87.9} & \textbf{69.2} & \textbf{67.4} & \textbf{9.1} & \textbf{22.6} & \textbf{33.1} \\

\bottomrule[1.2pt]
\end{tabular}}
\caption{Ablation study on the MemoryAgentBench and LongMemEval benchmarks under DeepSeek-V4-Flash.}
\label{ablation}
% \vspace{-4pt}
\end{table*}

\paragraph{Evaluation Metrics}
We adopt two complementary categories of metrics for evaluation. For \textbf{Task-Level Performance}, we follow the standard evaluation protocols of each benchmark. Specifically, on MemoryAgentBench, we report the performance on \textit{Fact Consolidation} tasks, including \textit{Single-Hop Fact Consolidation} (FC-SH), \textit{Multi-Hop Fact Consolidation} (FC-MH), and the overall score. On LongMemEval and LoCoMo, we report Exact Match (EM), F1, BERTScore, and judge-based evaluation scores.
For \textbf{Memory Correlation Bias}, we evaluate evidence arbitration under correlation-aware settings using four metrics: (\expandafter{\romannumeral1}) \textit{Correct Minority Recovery} (CMR), which measures the proportion of cases where the model recovers the correct answer when supporting evidence is outnumbered by correlated memories; (\expandafter{\romannumeral2}) \textit{Replication Sensitivity} (RS), which measures the impact of increasing redundant memories from the same evidence source on model decisions; (\expandafter{\romannumeral3}) \textit{Independent Evidence Gain} (IEG), which quantifies the benefit of incorporating additional independent evidence; and (\expandafter{\romannumeral4}) \textit{Evidence Resolution Rate} (ERR), which measures the proportion of evidence-insufficient instances that are correctly resolved.

\paragraph{Implementation Details}
We conduct our experiments with DeepSeek-V4-Flash~\citep{deepseekv4flash} and Qwen3.6-27B~\citep{qwen36blog} as the backbone LLMs. All baseline methods and \textsc{CAMA} use the same backbone models, memory pools, and retrieval settings for fair comparison. The evidence assignment module is trained on the training split and frozen during evaluation, while the recovery policy is optimized during training and fixed during inference. We set the number of retrieved memories to $K=10$, the number of latent evidence slots to $J=6$, and the recovery budget to $B=3$, respectively. For evidence estimation, we set the Hill diversity order to $\alpha=2$ and the provenance prior strength to $\mu=0.5$. For recovery policy optimization, we use a discount factor of $\gamma=0.95$, a value coefficient of $c_v=0.5$, and an entropy coefficient of $c_e=0.01$ by default.
% All experiments are repeated with random seeds, and the average results are reported.
% We conduct our experiments on Qwen3.5-27B~\citep{qwen35blog}, LLaMA-3.1-8B~\citep{Dubey2024TheL3, touvron2023llama}. The backbone LLMs generate both initial trajectories and reference-conditioned suffixes. We use the same preference-conflict construction for training and evaluation splits. The prefix consistency head and rollback policy are trained only on the training split and are frozen during evaluation. We default the number of rollback candidates to $M=2$, the rollback-drop coefficient to $\eta=1.0$, the intervention cost weight to $\lambda_c=0.5$, and the KL regularization weight to $\beta=1.0$. For policy training, grouped intervention outcomes are computed using the same evaluator, and the rollback policy is updated every $T_{\pi}=4$ rounds.

\begin{figure}[t]
\centering
\subfloat{
    \includegraphics[width=0.23\textwidth]{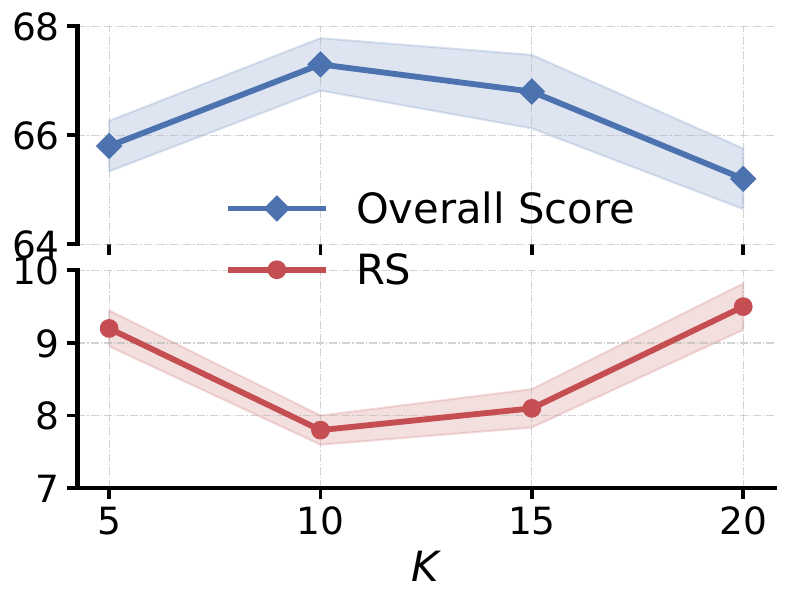}}
\subfloat{
    \includegraphics[width=0.23\textwidth]{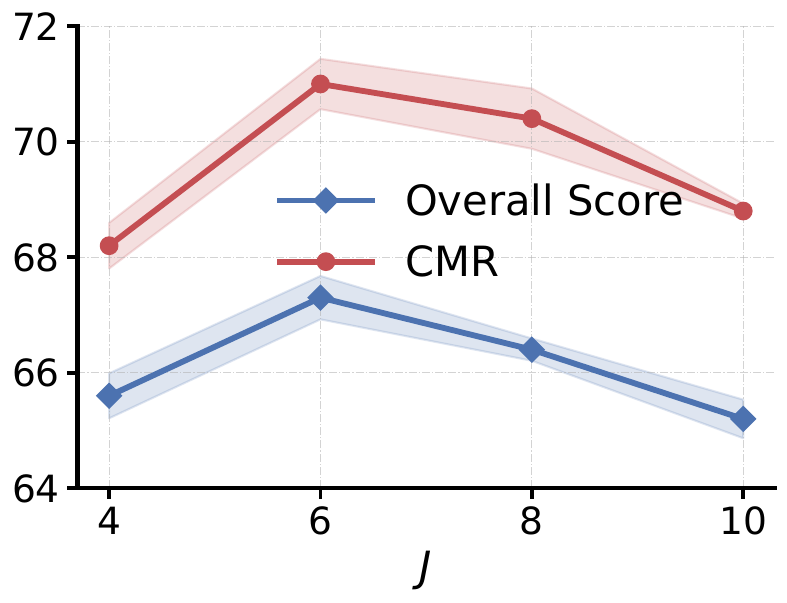}}
\vspace{5pt}
\subfloat{
    \includegraphics[width=0.23\textwidth]{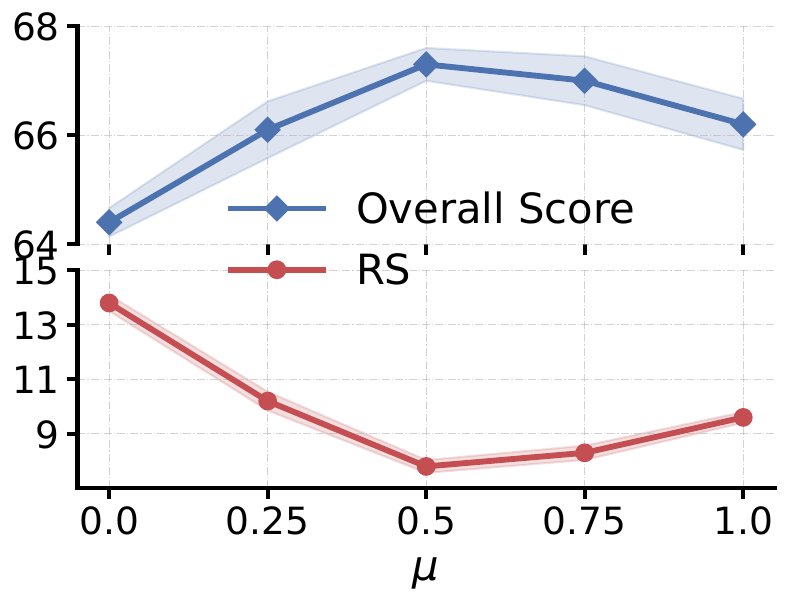}}
\subfloat{
    \includegraphics[width=0.23\textwidth]{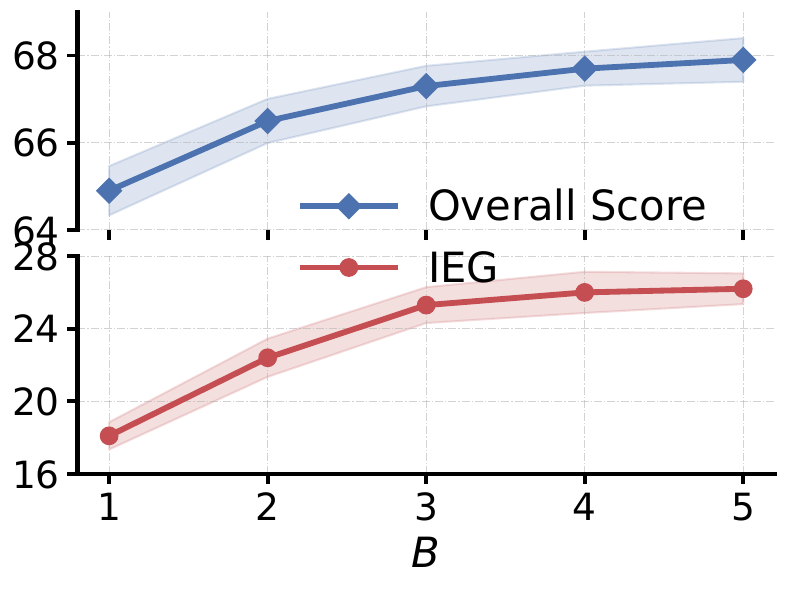}}
\caption{Hyperparameter sensitivity analysis on MemoryAgentBench with DeepSeek-V4-Flash.}
\label{hyper}
% \vspace{-6pt}
\end{figure}

\subsection{Experimental Results}
\paragraph{Overall Performance}
Table~\ref{accuracy} shows that \textsc{CAMA} consistently achieves superior performance across different benchmarks and backbones, demonstrating its effectiveness in long-term memory reasoning. Compared with retrieval-based methods that directly aggregate retrieved memories and memory management approaches that focus on memory organization, \textsc{CAMA} explicitly models latent evidence dependencies to avoid redundant memory amplification and identify reliable evidence. While multi-agent approaches improve reasoning through interaction and aggregation, \textsc{CAMA} further addresses memory correlation and missing evidence recovery, leading to more robust memory utilization.

Table~\ref{correlation_bias} further evaluates \textsc{CAMA} under correlation-aware settings. The consistent improvements across correlation metrics demonstrate that \textsc{CAMA} effectively identifies correlated memories and mitigates memory correlation bias. These gains stem from evidence-level decoupling and provenance modeling, which reduce redundant evidence interference, together with evidence recovery mechanisms that acquire missing evidence when needed.

\paragraph{Ablation Analysis}
Table~\ref{ablation} evaluates the contribution of each component in \textsc{CAMA}. Removing any component consistently degrades performance, confirming the necessity of evidence decoupling and adaptive recovery. Specifically, removing evidence decoupling or provenance modeling weakens correlation bias mitigation by failing to identify source-level dependencies among memories, while removing recovery components reduces the ability to acquire missing independent evidence. These results demonstrate that each component provides complementary benefits for reliable arbitration under correlated memory conditions.

\paragraph{Hyperparameter Sensitivity}
We analyze the sensitivity of \textsc{CAMA} to key hyperparameters, including the number of retrieved memories $K$, latent evidence slots $J$, provenance prior strength $\mu$, and budget $B$. As shown in Figure~\ref{hyper}, \textsc{CAMA} remains robust across different settings. Increasing $K$ improves evidence coverage but introduces redundancy when over-retrieved, validating the need for correlation-aware estimation. The choice of $J$ balances evidence factor separation and fragmentation, while a moderate $\mu$ balances provenance guidance and assignment flexibility. Increasing $B$ improves evidence recovery with diminishing returns. These results demonstrate that \textsc{CAMA} is robust to hyperparameter variation rather than depending on a narrowly tuned configuration.

\begin{table}[t]
\centering
\renewcommand\arraystretch{1.0}
\resizebox{0.48\textwidth}{!}{
\begin{tabular}{c|ccccc} 
\toprule[1.2pt]

\makecell{\centering\textbf{Methods}} & \makecell{Avg. \\ Latency} & \makecell{LLM \\ Calls} & \makecell{Token \\ Cost (k)} & \makecell{Context \\ Len (k)}  & \makecell{$\Delta$ Acc./ \\ kToken}  \\   \cmidrule[0.5pt](l{1pt}r{0pt}){1-6}

\textsc{Vanilla RAG} & 1.8 & 1.0 & 3.2 & 3.1 & -- \\

\textsc{Majority Voting} & 4.6 & 5.0 & 12.8 & 3.1 & 0.04 \\

\textsc{HippoRAG} & 3.2 & 2.0 & 5.9 & 4.4 & 1.08 \\

\textsc{MAD} & 9.8 & 8.4 & 24.3 & 9.6 & 0.39 \\

\textsc{MADAM-RAG} & 11.4 & 10.6 & 28.9 & 11.2 & 0.42 \\ \midrule[0.8pt]

\textsc{CAMA}(Ours) & 6.7 & 4.2 & 14.6 & 6.8 & 1.10 \\

\bottomrule[1.2pt]
\end{tabular}}
\caption{Efficiency analysis on the MemoryAgentBench benchmark under DeepSeek-V4-Flash.}
\label{Efficiency}
% \vspace{-7pt}
\end{table}

\paragraph{Efficiency Analysis}
We evaluate the efficiency from different perspectives. $\Delta$ Acc./kToken measures the accuracy improvement over \textsc{Vanilla RAG} normalized by the consumed thousands of tokens. Table~\ref{Efficiency} reports the efficiency comparison of \textsc{CAMA}. Although \textsc{CAMA} introduces additional costs for evidence modeling and adaptive recovery, it achieves a favorable accuracy--efficiency trade-off. By selectively activating recovery only when the retrieved evidence is insufficient, \textsc{CAMA} avoids unnecessary computation while maintaining effective evidence acquisition. The improved accuracy gain per token further indicates that the additional computation is primarily used for identifying independent evidence rather than repeatedly aggregating correlated memories. These results demonstrate the efficiency of \textsc{CAMA} in achieving reliable arbitration under correlated memory conditions.

\section{Conclusion}
In this paper, we identify \emph{Memory Correlation Bias} as a critical challenge in long-term multi-agent systems, where correlated memories may create false majorities and lead to persistent reasoning errors. We propose \textsc{CAMA}, a correlation-aware memory arbitration framework that evaluates evidence at the level of latent independent factors and selectively recovers missing evidence when current memories are insufficient. By combining neuro-symbolic evidence modeling, factor-level arbitration, and adaptive recovery, \textsc{CAMA} enables more reliable decisions under correlated memory conditions. Extensive experiments show that \textsc{CAMA} consistently outperforms existing approaches with favorable efficiency. Our work demonstrates that effective memory reasoning requires not only accumulating more memories, but also understanding their underlying dependencies.

% % \section*{Acknowledgments}
% % AAAI is especially grateful to Peter Patel Schneider for his work in implementing the original aaai.sty file, liberally using the ideas of other style hackers, including Barbara Beeton. We also acknowledge with thanks the work of George Ferguson for his guide to using the style and BibTeX files --- which has been incorporated into this document --- and Hans Guesgen, who provided several timely modifications, as well as the many others who have, from time to time, sent in suggestions on improvements to the AAAI style. We are especially grateful to Francisco Cruz, Marc Pujol-Gonzalez, and Mico Loretan for the improvements to the Bib\TeX{} and \LaTeX{} files made in 2020. \cite{hcrt:83}

% % The preparation of the \LaTeX{} and Bib\TeX{} files that implement these instructions was supported by Schlumberger Palo Alto Research, AT\&T Bell Laboratories, Morgan Kaufmann Publishers, The Live Oak Press, LLC, and AAAI Press. Bibliography style changes were added by Sunil Issar. \verb+\+pubnote was added by J. Scott Penberthy. George Ferguson added support for printing the AAAI copyright slug. Additional changes to aaai2027.sty and aaai2027.bst have been made by Francisco Cruz, Marc Pujol-Gonzalez, and Mico Loretan.

% % \bigskip
% % \noindent Thank you for reading these instructions carefully. We look forward to receiving your electronic files!

% \newpage

\bibliography{reference}

@misc{qwen36blog,
    title = {{Qwen3.6-27B}: Flagship-Level Coding in a 27B Dense Model},
    url = {https://qwen.ai/blog?id=qwen3.6-27b},
    author = {{Qwen Team}},
    month = {April},
    year = {2026}
}

@misc{deepseekv4flash,
    title = {DeepSeek V4 Preview Release},
    url = {https://api-docs.deepseek.com/news/news260424/},
    author = {DeepSeek},
    month = {April},
    year = {2026}
}

@article{hitzler2022neural,
  title={Neural-symbolic learning and reasoning: A survey and interpretation},
  author={Hitzler, P and Sarker, MK and Besold, TR and Garcez, AD and Bader, S and Bowman, H and Domingos, P and Hitzler, P and K{\"u}hnberger, KU and Lamb, LC and others},
  journal={Frontiers in artificial intelligence and applications},
  volume={342},
  pages={1--51},
  year={2022}
}

@inproceedings{garcez2015neural,
  title={Neural-Symbolic Learning and Reasoning: Contributions and Challenges.},
  author={Garcez, Artur S d'Avila and Besold, Tarek R and De Raedt, Luc and F{\"o}ldiak, Peter and Hitzler, Pascal and Icard, Thomas and K{\"u}hnberger, Kai-Uwe and Lamb, Luis C and Miikkulainen, Risto and Silver, Daniel L},
  booktitle={AAAI Spring Symposia},
  pages={18--21},
  year={2015}
}

@inproceedings{hu2026evaluating,
title={Evaluating Memory in {LLM} Agents via Incremental Multi-Turn Interactions},
author={Yuanzhe Hu and Yu Wang and Julian McAuley},
booktitle={The Fourteenth International Conference on Learning Representations},
year={2026},
url={https://openreview.net/forum?id=DT7JyQC3MR}
}

@inproceedings{maharana2024evaluating,
  title={Evaluating very long-term conversational memory of llm agents},
  author={Maharana, Adyasha and Lee, Dong-Ho and Tulyakov, Sergey and Bansal, Mohit and Barbieri, Francesco and Fang, Yuwei},
  booktitle={Proceedings of the 62nd Annual Meeting of the Association for Computational Linguistics (Volume 1: Long Papers)},
  pages={13851--13870},
  year={2024}
}

@inproceedings{wu2025longmemeval,
title={LongMemEval: Benchmarking Chat Assistants on Long-Term Interactive Memory},
author={Di Wu and Hongwei Wang and Wenhao Yu and Yuwei Zhang and Kai-Wei Chang and Dong Yu},
booktitle={The Thirteenth International Conference on Learning Representations},
year={2025},
url={https://openreview.net/forum?id=pZiyCaVuti}
}

@article{lewis2020retrieval,
  title={Retrieval-augmented generation for knowledge-intensive nlp tasks},
  author={Lewis, Patrick and Perez, Ethan and Piktus, Aleksandra and Petroni, Fabio and Karpukhin, Vladimir and Goyal, Naman and K{\"u}ttler, Heinrich and Lewis, Mike and Yih, Wen-tau and Rockt{\"a}schel, Tim and others},
  journal={Advances in neural information processing systems},
  volume={33},
  pages={9459--9474},
  year={2020}
}

@inproceedings{
wang2023selfconsistency,
title={Self-Consistency Improves Chain of Thought Reasoning in Language Models},
author={Xuezhi Wang and Jason Wei and Dale Schuurmans and Quoc V Le and Ed H. Chi and Sharan Narang and Aakanksha Chowdhery and Denny Zhou},
booktitle={The Eleventh International Conference on Learning Representations },
year={2023},
url={https://openreview.net/forum?id=1PL1NIMMrw}
}

@article{gutierrez2024hipporag,
  title={Hipporag: Neurobiologically inspired long-term memory for large language models},
  author={Guti{\'e}rrez, Bernal J and Shu, Yiheng and Gu, Yu and Yasunaga, Michihiro and Su, Yu},
  journal={Advances in neural information processing systems},
  volume={37},
  pages={59532--59569},
  year={2024}
}

@article{chhikara2025mem0,
  title={Mem0: Building production-ready ai agents with scalable long-term memory},
  author={Chhikara, Prateek and Khant, Dev and Aryan, Saket and Singh, Taranjeet and Yadav, Deshraj},
  journal={arXiv preprint arXiv:2504.19413},
  year={2025}
}

@inproceedings{liang2024encouraging,
  title={Encouraging divergent thinking in large language models through multi-agent debate},
  author={Liang, Tian and He, Zhiwei and Jiao, Wenxiang and Wang, Xing and Wang, Yan and Wang, Rui and Yang, Yujiu and Shi, Shuming and Tu, Zhaopeng},
  booktitle={Proceedings of the 2024 conference on empirical methods in natural language processing},
  pages={17889--17904},
  year={2024}
}

@article{wang2025retrieval,
  title={Retrieval-augmented generation with conflicting evidence},
  author={Wang, Han and Prasad, Archiki and Stengel-Eskin, Elias and Bansal, Mohit},
  journal={arXiv preprint arXiv:2504.13079},
  year={2025}
}

@article{rezazadeh2025collaborative,
  title={Collaborative memory: Multi-user memory sharing in llm agents with dynamic access control},
  author={Rezazadeh, Alireza and Li, Zichao and Lou, Ange and Zhao, Yuying and Wei, Wei and Bao, Yujia},
  journal={arXiv preprint arXiv:2505.18279},
  year={2025}
}

@article{zhang2025g,
  title={G-memory: Tracing hierarchical memory for multi-agent systems},
  author={Zhang, Guibin and Fu, Muxin and Wang, Kun and Wan, Frank and Yu, Miao and Yan, Shuicheng},
  journal={Advances in Neural Information Processing Systems},
  volume={38},
  pages={12988--13018},
  year={2025}
}

@article{yang2026auditing,
  title={Auditing multi-agent llm reasoning trees outperforms majority vote and llm-as-judge},
  author={Yang, Wei and Li, Shixuan and Ping, Heng and Zhang, Peiyu and Bogdan, Paul and Thomason, Jesse},
  journal={arXiv preprint arXiv:2602.09341},
  year={2026}
}

@inproceedings{ai2026beyond,
title={Beyond Majority Voting: {LLM} Aggregation by Leveraging Higher-Order Information},
author={Rui Ai and Yuqi Pan and David Simchi-Levi and Milind Tambe and Haifeng Xu},
booktitle={Forty-third International Conference on Machine Learning},
year={2026},
url={https://openreview.net/forum?id=ZVyd4r9Xl5}
}

@article{kohli2026nine,
  title={Nine Judges, Two Effective Votes: Correlated Errors Undermine LLM Evaluation Panels},
  author={Kohli, Guneet},
  journal={arXiv preprint arXiv:2605.29800},
  year={2026}
}

@inproceedings{kang2025memory,
  title={Memory os of ai agent},
  author={Kang, Jiazheng and Ji, Mingming and Zhao, Zhe and Bai, Ting},
  booktitle={Proceedings of the 2025 Conference on Empirical Methods in Natural Language Processing},
  pages={25972--25981},
  year={2025}
}

@inproceedings{yu2026agentic,
    title = "Agentic Memory: Learning Unified Long-Term and Short-Term Memory Management for Large Language Model Agents",
    author = "Yu, Yi  and
      Yao, Liuyi  and
      Xie, Yuexiang  and
      Tan, Qingquan  and
      Feng, Jiaqi  and
      Li, Yaliang  and
      Wu, Libing",
    editor = "Liakata, Maria  and
      Moreira, Viviane P.  and
      Zhang, Jiajun  and
      Jurgens, David",
    booktitle = "Proceedings of the 64th Annual Meeting of the {A}ssociation for {C}omputational {L}inguistics (Volume 1: Long Papers)",
    month = jul,
    year = "2026",
    address = "San Diego, California, United States",
    publisher = "Association for Computational Linguistics",
    url = "https://aclanthology.org/2026.acl-long.981/",
    doi = "10.18653/v1/2026.acl-long.981",
    pages = "21457--21483",
    ISBN = "979-8-89176-390-6",
}

@article{liu2026consensus,
  title={The Consensus Trap: Rescuing Multi-Agent LLMs from Adversarial Majorities via Token-Level Collaboration},
  author={Liu, Jiayuan and Du, Shiyi and Du, Weihua and Guo, Mingyu and Conitzer, Vincent},
  journal={arXiv preprint arXiv:2604.17139},
  year={2026}
}

@inproceedings{tan2025prospect,
  title={In prospect and retrospect: Reflective memory management for long-term personalized dialogue agents},
  author={Tan, Zhen and Yan, Jun and Hsu, I-Hung and Han, Rujun and Wang, Zifeng and Le, Long and Song, Yiwen and Chen, Yanfei and Palangi, Hamid and Lee, George and others},
  booktitle={Proceedings of the 63rd Annual Meeting of the Association for Computational Linguistics (Volume 1: Long Papers)},
  pages={8416--8439},
  year={2025}
}

@inproceedings{salama2025meminsight,
  title={Meminsight: Autonomous memory augmentation for llm agents},
  author={Salama, Rana and Cai, Jason and Yuan, Michelle and Currey, Anna and Sunkara, Monica and Zhang, Yi and Benajiba, Yassine},
  booktitle={Proceedings of the 2025 Conference on Empirical Methods in Natural Language Processing},
  pages={33124--33140},
  year={2025}
}

@inproceedings{huang2026ama,
    title = "{AMA}: Adaptive Memory via Multi-Agent Collaboration",
    author = "Huang, Weiquan  and
      Wang, Zixuan  and
      Lin, Hehai  and
      Wang, Sudong  and
      Xu, Bo  and
      Li, Qian  and
      Zhu, Beier  and
      Yang, Linyi  and
      Qin, Chengwei",
    editor = "Liakata, Maria  and
      Moreira, Viviane P.  and
      Zhang, Jiajun  and
      Jurgens, David",
    booktitle = "Findings of the {A}ssociation for {C}omputational {L}inguistics: {ACL} 2026",
    month = jul,
    year = "2026",
    address = "San Diego, California, United States",
    publisher = "Association for Computational Linguistics",
    url = "https://aclanthology.org/2026.findings-acl.152/",
    doi = "10.18653/v1/2026.findings-acl.152",
    pages = "3099--3120",
    ISBN = "979-8-89176-395-1",
}

@article{zhang2025survey,
  title={A survey on the memory mechanism of large language model-based agents},
  author={Zhang, Zeyu and Dai, Quanyu and Bo, Xiaohe and Ma, Chen and Li, Rui and Chen, Xu and Zhu, Jieming and Dong, Zhenhua and Wen, Ji-Rong},
  journal={ACM Transactions on Information Systems},
  volume={43},
  number={6},
  pages={1--47},
  year={2025},
  publisher={ACM New York, NY}
}

@article{xu2025mem,
  title={A-mem: Agentic memory for llm agents},
  author={Xu, Wujiang and Liang, Zujie and Mei, Kai and Gao, Hang and Tan, Juntao and Zhang, Yongfeng},
  journal={Advances in Neural Information Processing Systems},
  volume={38},
  pages={17577--17604},
  year={2025}
}

@inproceedings{xu2026chain,
    title = "Chain-of-Memory: Lightweight Memory Construction with Dynamic Evolution for {LLM} Agents",
    author = "Xu, Xiucheng  and
      Xu, Bingbing  and
      Xueyun, Tian  and
      Huang, Zihe  and
      Chen, Rongxin  and
      Yunfan, Li  and
      Shen, Huawei",
    editor = "Liakata, Maria  and
      Moreira, Viviane P.  and
      Zhang, Jiajun  and
      Jurgens, David",
    booktitle = "Proceedings of the 64th Annual Meeting of the {A}ssociation for {C}omputational {L}inguistics (Volume 1: Long Papers)",
    month = jul,
    year = "2026",
    address = "San Diego, California, United States",
    publisher = "Association for Computational Linguistics",
    url = "https://aclanthology.org/2026.acl-long.534/",
    doi = "10.18653/v1/2026.acl-long.534",
    pages = "11618--11631",
    ISBN = "979-8-89176-390-6",
}

@article{du2025memr,
  title={MemR $^3$: Memory Retrieval via Reflective Reasoning for LLM Agents},
  author={Du, Xingbo and Li, Loka and Zhang, Duzhen and Song, Le},
  journal={arXiv preprint arXiv:2512.20237},
  year={2025}
}

@inproceedings{souza2025prov,
  title={PROV-AGENT: Unified provenance for tracking AI agent interactions in agentic workflows},
  author={Souza, Renan and Gueroudji, Amal and DeWitt, Stephen and Rosendo, Daniel and Ghosal, Tirthankar and Ross, Robert and Balaprakash, Prasanna and Da Silva, Rafael Ferreira},
  booktitle={2025 IEEE International Conference on eScience (eScience)},
  pages={467--473},
  year={2025},
  organization={IEEE}
}

@article{wang2026agent,
  title={From Agent Traces to Trust: A Survey of Evidence Tracing and Execution Provenance in LLM Agents},
  author={Wang, Yiqi and Zhang, Jiaqi and Cai, Taotao and Liu, Zirui and Sun, Qingqiang and Sun, Zequn and Wu, Zhangkai and Dong, Manqing and Zheng, Mingkai and Yin, Xuefei and others},
  journal={arXiv preprint arXiv:2606.04990},
  year={2026}
}

@article{lu2026mma,
  title={Mma: Multimodal memory agent},
  author={Lu, Yihao and Cheng, Wanru and Zhang, Zeyu and Tang, Hao},
  journal={arXiv preprint arXiv:2602.16493},
  year={2026}
}

@article{liang2025cognitive,
  title={Cognitive-inspired xLSTM for multi-agent information retrieval},
  author={Liang, Li and Wang, Huan and Wang, Kai},
  journal={Scientific Reports},
  volume={15},
  number={1},
  pages={36121},
  year={2025},
  publisher={Nature Publishing Group UK London}
}

@inproceedings{taubenfeld2025confidence,
  title={Confidence improves self-consistency in llms},
  author={Taubenfeld, Amir and Sheffer, Tom and Ofek, Eran and Feder, Amir and Goldstein, Ariel and Gekhman, Zorik and Yona, Gal},
  booktitle={Findings of the Association for Computational Linguistics: ACL 2025},
  pages={20090--20111},
  year={2025}
}

@inproceedings{hwang2025retrieval,
    title = "Retrieval-Augmented Generation with Estimation of Source Reliability",
    author = "Hwang, Jeongyeon  and
      Park, Junyoung  and
      Park, Hyejin  and
      Kim, Dongwoo  and
      Park, Sangdon  and
      Ok, Jungseul",
    editor = "Christodoulopoulos, Christos  and
      Chakraborty, Tanmoy  and
      Rose, Carolyn  and
      Peng, Violet",
    booktitle = "Proceedings of the 2025 Conference on Empirical Methods in Natural Language Processing",
    month = nov,
    year = "2025",
    address = "Suzhou, China",
    publisher = "Association for Computational Linguistics",
    url = "https://aclanthology.org/2025.emnlp-main.1738/",
    doi = "10.18653/v1/2025.emnlp-main.1738",
    pages = "34279--34303",
    ISBN = "979-8-89176-332-6",
}

@inproceedings{razghandi2025cer,
  title={Cer: Confidence enhanced reasoning in llms},
  author={Razghandi, Ali and Hosseini, Seyed Mohammad Hadi and Baghshah, Mahdieh Soleymani},
  booktitle={Proceedings of the 63rd Annual Meeting of the Association for Computational Linguistics (Volume 1: Long Papers)},
  pages={7918--7938},
  year={2025}
}

@inproceedings{kim2025correlated,
  title={Correlated Errors in Large Language Models},
  author={Kim, Elliot Myunghoon and Garg, Avi and Peng, Kenny and Garg, Nikhil},
  booktitle={International Conference on Machine Learning},
  pages={30038--30066},
  year={2025},
  organization={PMLR}
}

@inproceedings{ge2025resolving,
  title={Resolving conflicting evidence in automated fact-checking: a study on retrieval-augmented LLMs},
  author={Ge, Ziyu and Wu, Yuhao and Chin, Daniel Wai Kit and Lee, Roy Ka-Wei and Cao, Rui},
  booktitle={Proceedings of the Thirty-Fourth International Joint Conference on Artificial Intelligence},
  pages={9656--9664},
  year={2025}
}

@article{peer2025ata,
  title={ATA: A Neuro-Symbolic Approach to Implement Autonomous and Trustworthy Agents},
  author={Peer, David and Stabinger, Sebastian},
  journal={arXiv preprint arXiv:2510.16381},
  year={2025}
}

@inproceedings{yang2025neuro,
  title={Neuro-symbolic artificial intelligence: towards improving the reasoning abilities of large language models},
  author={Yang, Xiao-Wen and Shao, Jie-Jing and Guo, Lan-Zhe and Zhang, Bo-Wen and Zhou, Zhi and Jia, Lin-Han and Dai, Wang-Zhou and Li, Yu-Feng},
  booktitle={Proceedings of the Thirty-Fourth International Joint Conference on Artificial Intelligence},
  pages={10770--10778},
  year={2025}
}

@inproceedings{chang2025main,
    title = "{MAIN}-{RAG}: Multi-Agent Filtering Retrieval-Augmented Generation",
    author = "Chang, Chia-Yuan  and
      Jiang, Zhimeng  and
      Rakesh, Vineeth  and
      Pan, Menghai  and
      Yeh, Chin-Chia Michael  and
      Wang, Guanchu  and
      Hu, Mingzhi  and
      Xu, Zhichao  and
      Zheng, Yan  and
      Das, Mahashweta  and
      Zou, Na",
    editor = "Che, Wanxiang  and
      Nabende, Joyce  and
      Shutova, Ekaterina  and
      Pilehvar, Mohammad Taher",
    booktitle = "Proceedings of the 63rd Annual Meeting of the Association for Computational Linguistics (Volume 1: Long Papers)",
    month = jul,
    year = "2025",
    address = "Vienna, Austria",
    publisher = "Association for Computational Linguistics",
    url = "https://aclanthology.org/2025.acl-long.131/",
    doi = "10.18653/v1/2025.acl-long.131",
    pages = "2607--2622",
    ISBN = "979-8-89176-251-0",
}

@inproceedings{peng2025cafe,
  title={CAFE: Retrieval Head-based Coarse-to-Fine Information Seeking to Enhance Multi-Document QA Capability},
  author={Peng, Han and Jiang, Jinhao and Dong, Zican and Zhao, Wayne Xin and Fang, Lei},
  booktitle={Proceedings of the 2025 Conference on Empirical Methods in Natural Language Processing},
  pages={12966--12978},
  year={2025}
}

@inproceedings{lin2025rje,
  title={RJE: A Retrieval-Judgment-Exploration Framework for Efficient Knowledge Graph Question Answering with LLMs},
  author={Lin, Can and Jiang, Zhengwang and Zheng, Ling and Zhao, Qi and Zhang, Yuhang and Song, Qi and Zhou, Wangqiu},
  booktitle={Proceedings of the 2025 Conference on Empirical Methods in Natural Language Processing},
  pages={17288--17305},
  year={2025}
}

@inproceedings{tran2025rare,
    title = "{RARE}: Retrieval-Augmented Reasoning Enhancement for Large Language Models",
    author = "Tran, Hieu  and
      Yao, Zonghai  and
      Yang, Zhichao  and
      Wang, Junda  and
      Zhang, Yifan  and
      Han, Shuo  and
      Feiyun Ouyang  and
      Yu, Hong",
    editor = "Che, Wanxiang  and
      Nabende, Joyce  and
      Shutova, Ekaterina  and
      Pilehvar, Mohammad Taher",
    booktitle = "Proceedings of the 63rd Annual Meeting of the Association for Computational Linguistics (Volume 1: Long Papers)",
    month = jul,
    year = "2025",
    address = "Vienna, Austria",
    publisher = "Association for Computational Linguistics",
    url = "https://aclanthology.org/2025.acl-long.896/",
    doi = "10.18653/v1/2025.acl-long.896",
    pages = "18305--18330",
    ISBN = "979-8-89176-251-0",
}

% % Check whether the conference requires a reproducibility checklist to be included in the paper.
% % If so, you can uncomment the following line and ajust the path to include it.
% % \input{ReproducibilityChecklist.tex}

\newpage

\appendix

\section{Algorithm}

\subsection{Inference Procedure and Complexity} \label{sec:inference}
Algorithm~\ref{alg:cama} summarizes the inference procedure. At each recovery step, CAMA updates the candidate hypotheses, evidence assignments, and arbitration posterior based on the current memory slice and provenance structure. The procedure terminates when the recovery budget is exhausted, the evidence-sufficiency criterion is satisfied, or the learned policy selects \textsc{Stop}. After each nonterminal recovery action, the updated memory slice and provenance structure are used to recompute the latent evidence factors and arbitration results. The final output contains the selected
hypothesis together with its factor-level attribution.

For a current slice of size $K_t$, the set encoder incurs $O(K_t^2)$ complexity due to self-attention. With a recovery budget $B$, the total encoding cost is $O(\sum_{t=0}^{B}K_t^2)$, which depends on the query-local memory slice rather than the full memory store. Additional retrieval cost is introduced only by \textsc{Expand}, while \textsc{Trace} follows existing provenance links. When the initial retrieval already provides sufficient evidence and confident arbitration, the process terminates after a single-pass factor-level arbitration.

\subsection{Correlation-Aware Evaluation Metrics}
\label{app:metrics}
To quantify memory correlation bias, we define four correlation-aware metrics based on the constructed evaluation instances. Let $\mathcal{D}$ denote the set of evaluation cases. Each case contains a set of retrieved memories $\mathcal{M}$, where memories may originate from either the same underlying evidence source (correlated memories) or distinct sources (independent evidence). We denote the model prediction before and after applying \textsc{CAMA} as $y$ and $y^{*}$, respectively, and use $\mathbb{I}(\cdot)$ as the indicator function.

% \begin{algorithm}[ht]
% \caption{Correlation-Aware Memory Arbitration}
% \label{alg:cama}
% \begin{algorithmic}[1]
% \REQUIRE Query $q$, memory store $\mathcal{M}$, retrieval interface, budget $B$
% \ENSURE Conclusion $\hat h$ and cluster-level attribution
% \STATE $\mathcal{C}^{(0)}_q\gets\operatorname{Retrieve}(q,\mathcal{M};K)$
% \STATE $G^{(0)}_{\mathrm{prov}}\gets$ observed local relations in $\mathcal{C}^{(0)}_q$
% \FOR{$t=0,\ldots,B$}
%     \STATE Update $\mathcal{H}^{(t)}_q$ from $\mathcal{C}^{(t)}_q$
%     \STATE $Z^{(t)}\gets f_{\Theta}(q,\mathcal{C}^{(t)}_q,G^{(t)}_{\mathrm{prov}})$
%     \STATE Compute $e^{(t)}$, $N^{(t)}_{\mathrm{eff}}$, and $\boldsymbol\kappa^{(t)}$
%     \STATE Compute $P^{(t)}$ using Equations~\eqref{eq:cluster-support}--\eqref{eq:arbitration-posterior}
%     \IF{$t=B$}
%         \STATE \textbf{break}
%     \ENDIF
%     \STATE $A_t\sim\pi_{\omega}(\cdot\mid S_t)$
%     \IF{$A_t=\textsc{Stop}$}
%         \STATE \textbf{break}
%     \ELSIF{$A_t=\textsc{Expand}(q')$}
%         \STATE Update $\mathcal{C}^{(t+1)}_q$ and $G^{(t+1)}_{\mathrm{prov}}\gets G^{(t)}_{\mathrm{prov}}$
%     \ELSIF{$A_t=\textsc{Trace}(m_i)$}
%         \STATE Recover $m_p$ and update $\mathcal{C}^{(t+1)}_q,G^{(t+1)}_{\mathrm{prov}}$
%     \ENDIF
% \ENDFOR
% \STATE $\hat h\gets\arg\max_{h\in\mathcal{H}^{(t)}_q}P^{(t)}(h)$
% \STATE \textbf{return} $\hat h$ and cluster-level attribution
% \end{algorithmic}
% \end{algorithm}

\begin{algorithm}[t]
\caption{Correlation-Aware Memory Arbitration}
\label{alg:cama}
\begin{algorithmic}[1]
\REQUIRE Query $q$, memory store $\mathcal{M}$, retrieval sizes
$K,K_{\mathrm{add}}$, budget $B$, thresholds $\tau_N,\tau_H$
\ENSURE Conclusion $\hat h$ and factor-level attribution

\STATE $\mathcal{C}^{(0)}_q
\gets \operatorname{Retrieve}(q,\mathcal{M};K)$
\STATE $G^{(0)}_{\mathrm{prov}}
\gets \operatorname{LocalProv}(\mathcal{C}^{(0)}_q)$

\FOR{$t=0,\ldots,B$}
    \STATE Infer $\mathcal{H}^{(t)}_q$, $Z^{(t)}$ and compute
    $e^{(t)}$, $N^{(t)}_{\mathrm{eff}}$,
    $\boldsymbol{\kappa}^{(t)}$, $\beta^{(t)}$,
    $\rho^{(t)}$, and $P^{(t)}$

    \STATE $\mathrm{sufficient}^{(t)}
    \gets
    (N^{(t)}_{\mathrm{eff}}\geq\tau_N)
    \land
    (\mathrm{H}(P^{(t)})\leq\tau_H)$

    \IF{$t=B$}
        \STATE \textbf{break}
    \ENDIF

    \STATE Generate expansion queries $\mathcal{Q}^{(t)}$ from $S_t$

    \STATE $\mathcal{A}^{(t)}
    \gets
    \{\textsc{Expand}(q'):q'\in\mathcal{Q}^{(t)}\}
    \cup
    \{\textsc{Trace}(m_i):m_i\text{ has a traceable parent}\}$

    \IF{$\mathrm{sufficient}^{(t)}$}
        \STATE $\mathcal{A}^{(t)}
        \gets\mathcal{A}^{(t)}\cup\{\textsc{Stop}\}$
    \ENDIF

    \STATE $A_t
    \sim
    \pi_{\omega}(\cdot\mid S_t,\mathcal{A}^{(t)})$

    \IF{$A_t=\textsc{Stop}$}
        \STATE \textbf{break}

    \ELSIF{$A_t=\textsc{Expand}(q')$}
        \STATE $\Delta\mathcal{C}^{(t)}
        \gets
        \operatorname{Retrieve}(q',\mathcal{M};K_{\mathrm{add}})$
        \STATE $\mathcal{C}^{(t+1)}_q
        \gets
        \mathcal{C}^{(t)}_q\cup\Delta\mathcal{C}^{(t)}$
        \STATE $G^{(t+1)}_{\mathrm{prov}}
        \gets
        G^{(t)}_{\mathrm{prov}}
        \cup
        \operatorname{LocalProv}
        (\Delta\mathcal{C}^{(t)},\mathcal{C}^{(t+1)}_q)$

    \ELSIF{$A_t=\textsc{Trace}(m_i)$}
        \STATE Recover the parent $m_p$ and its derivation edges
        $\mathcal{E}^{(t)}_p$
        \STATE $\mathcal{C}^{(t+1)}_q
        \gets
        \mathcal{C}^{(t)}_q\cup\{m_p\}$
        \STATE $G^{(t+1)}_{\mathrm{prov}}
        \gets
        G^{(t)}_{\mathrm{prov}}\cup\mathcal{E}^{(t)}_p$
    \ENDIF
\ENDFOR

\STATE $\hat h
\gets
\arg\max_{h\in\mathcal{H}^{(t)}_q}P^{(t)}(h)$
\STATE Construct factor-level attribution from the final evidence state
\STATE \textbf{return} $\hat h$ and factor-level attribution

\end{algorithmic}
\end{algorithm}

\section{Query-conditioned Evidence Decoupling}

The objective of evidence decoupling is to identify latent evidential factors underlying retrieved memories, rather than directly treating each memory entry as an independent evidence source. In \textsc{CAMA}, multiple memories may correspond to the same latent factor when they originate from shared observations, propagated summaries, or correlated reasoning processes. Given the retrieved memory slice $\mathcal{C}^{(t)}_q$, \textsc{CAMA} infers a soft assignment matrix $Z^{(t)}$ between memories and latent evidence factors. Each row of $Z^{(t)}$ represents the contribution distribution of a memory over different factors, while each column corresponds to a query-dependent evidential factor. These factors represent evidence units that support or contradict candidate hypotheses. Therefore, multiple memories assigned to the same factor are treated as redundant evidence, whereas memories associated with different factors provide potentially independent support. The dependency among memories is modeled in a query-conditioned manner through the overlap of their latent factor assignments:
\begin{align}
r^{(t)}_{ij}
=
\left\langle
\mathbf{z}^{(t)}_i,
\mathbf{z}^{(t)}_j
\right\rangle ,
\end{align}
where larger values indicate stronger evidential redundancy under the current query. Unlike static dependency based on memory metadata or agent identity, this formulation captures query-dependent evidence relationships. When provenance information is available, \textsc{CAMA} incorporates it as an auxiliary structural prior to guide evidence assignment without enforcing hard dependency constraints. The resulting factor-level representation is then used for effective independent evidence estimation and subsequent arbitration.

\section{Active Independent-Evidence Recovery}

The initial retrieval view may be insufficient for reliable arbitration because it only provides a partial observation of the underlying evidence structure. Specifically, the current memory slice may either over-represent existing evidence factors due to hidden correlations or fail to cover critical independent evidence factors required for decision making. Therefore, \textsc{CAMA} performs active evidence recovery to iteratively refine the retrieved evidence structure before final arbitration. At each recovery step, \textsc{CAMA} updates the current memory slice and re-estimates the latent evidence structure, including evidence diversity, assignment confidence, and arbitration uncertainty. The recovery policy selects an action according to whether the current evidence state requires additional evidence acquisition, dependency investigation, or termination. The recovery actions correspond to different types of evidence refinement. \textsc{Expand} discovers missing independent evidence factors by exploring alternative retrieval views. \textsc{Trace} identifies hidden dependencies among existing memories by following provenance relations. \textsc{Stop} terminates recovery when the current evidence structure provides sufficient independent support and the arbitration result is reliable. After each recovery action, the updated memory slice is used to re-estimate the evidence structure and perform subsequent arbitration.

\section{Detailed Dataset Descriptions}
We evaluate \textsc{CAMA} on three representative long-term memory benchmarks, covering different aspects of memory-augmented LLM agents, ranging from dynamic memory management under evolving interactions to long-horizon retrieval and reasoning over historical conversations.

\textbf{MemoryAgentBench}~\citep{hu2026evaluating} is designed to evaluate the memory capabilities of LLM-based agents through incremental multi-turn interactions. Different from conventional retrieval benchmarks that mainly measure whether relevant information can be retrieved from a static memory pool, MemoryAgentBench focuses on the dynamic maintenance of agent memories over time. It evaluates whether agents can effectively update, retrieve, and manage memories as new interactions accumulate, including scenarios involving conflicting memory updates and selective forgetting. Such settings naturally introduce situations where historical memories may become outdated, redundant, or inconsistent, making MemoryAgentBench suitable for evaluating whether an agent can identify reliable evidence among potentially correlated memories.

\textbf{LongMemEval}~\citep{wu2025longmemeval} evaluates long-term memory retrieval and reasoning capabilities of conversational agents under extended interaction histories. The benchmark contains long multi-session conversations where relevant information is distributed across historical interactions and requires agents to retrieve, integrate, and reason over long-term user memories. Compared with short-context dialogue benchmarks, LongMemEval emphasizes the ability to utilize accumulated user information under evolving conversational contexts. Since repeated interactions may produce multiple memory entries describing similar user states or historical events, LongMemEval provides a challenging setting for studying whether memory methods can distinguish independent evidence from correlated memory traces.

\textbf{LoCoMo}~\citep{maharana2024evaluating} focuses on long-term conversational memory reasoning over multi-session dialogues with evolving user states and historical interactions. The benchmark requires agents to answer queries by reasoning over information accumulated across multiple conversation sessions, including historical facts, temporal events, and user-related information. Due to the longitudinal nature of conversations, the benchmark contains naturally occurring memory dependencies where multiple records may originate from the same underlying event or user state. Therefore, LoCoMo serves as an effective testbed for evaluating correlation-aware memory arbitration in long-term conversational agents.

To specifically evaluate memory correlation bias, we further construct correlation-aware variants from the original benchmark instances. For each instance, we augment the memory pool with additional correlated memories derived from shared evidence sources as well as independent memories from distinct sources. Correlated memories are generated through controlled derivations, including paraphrasing and summarization, while preserving their original semantics and recording provenance relations between derived entries and their source memories. The original ground-truth answers remain unchanged, ensuring that the evaluation focuses on whether an agent can correctly identify independent evidence rather than relying on the quantity of retrieved memories. These correlation-aware variants provide a controlled evaluation environment for measuring the ability of \textsc{CAMA} to mitigate false majorities induced by correlated memories.

\paragraph{Correlation-aware Benchmark Construction}
Existing long-term memory benchmarks primarily evaluate retrieval and reasoning capabilities, but they do not explicitly measure the impact of correlated memories on evidence aggregation.
To evaluate memory correlation bias, we construct correlation-aware variants from the original benchmark instances while preserving their original task objectives and ground-truth answers.

Given an original memory pool $\mathcal{C}_q$, we augment it with two types of additional memories:
(1) \textit{correlated memories} derived from existing evidence sources, and
(2) \textit{independent memories} providing distinct query-relevant evidential factors.

For correlated memories, we generate additional memory entries by applying controlled transformations to existing memories, including paraphrasing and summarization.
These transformations preserve the underlying evidence while introducing surface-level diversity, simulating realistic scenarios where multiple agents or memory-writing processes record overlapping information from the same source.
The provenance relationship between each generated memory and its original source is explicitly recorded.

For independent memories, we introduce additional entries that provide complementary, non-overlapping evidence relevant to the query. These entries are drawn from distinct sources and treated as independent with respect to the current query because they correspond to different underlying evidential factors. The constructed memory pool can therefore be represented as:
\[
\widetilde{\mathcal{C}}_q
=
\mathcal{C}_q
\cup
\mathcal{C}^{\mathrm{corr}}_q
\cup
\mathcal{C}^{\mathrm{ind}}_q,
\]
where $\mathcal{C}^{\mathrm{corr}}_q$ denotes correlated memory entries generated from shared evidence sources, and $\mathcal{C}^{\mathrm{ind}}_q$ denotes entries providing distinct query-relevant evidential factors.

Importantly, the construction process does not modify the original queries or ground-truth answers. Instead, it preserves the target answer while perturbing the composition and multiplicity of the available evidence. This controlled construction enables us to evaluate how memory aggregation methods respond to correlated evidence and changes in evidence-source diversity.
The recorded provenance information further provides \textsc{CAMA} with structural priors for modeling potential shared-source dependencies and performing correlation-aware arbitration.

\section{Comparison Baselines}

We compare \textsc{CAMA} with representative baselines covering conventional retrieval-based aggregation, long-term memory management, and multi-agent reasoning approaches.

\begin{itemize}
\item \textsc{Vanilla RAG}~\citep{lewis2020retrieval} represents a standard retrieval-augmented generation pipeline, where retrieved memories are directly provided as additional context for generation. It treats all retrieved memory entries as independent information sources without explicitly modeling their dependencies or reliability, serving as a fundamental baseline for evaluating the impact of correlation-aware evidence arbitration.

\item \textsc{Majority Voting}~\citep{wang2023selfconsistency} represents consensus-based evidence aggregation strategies. 
It aggregates retrieved memories by selecting the hypothesis supported by the majority of memory entries, implicitly assuming that each memory provides an independent vote. Although effective when evidence sources are independent, such strategies may suffer from false majorities when multiple memories originate from the same underlying evidence source.

\item \textsc{Mem0}~\citep{chhikara2025mem0} is a long-term memory management framework that extracts salient information from interactions and maintains a compact memory store through memory addition, updating, and consolidation. Unlike direct retrieval methods, Mem0 focuses on scalable memory organization and adaptive memory maintenance for long-running LLM agents. We include Mem0 to evaluate whether existing memory management strategies can mitigate correlation issues through memory consolidation.

\item \textsc{HippoRAG}~\citep{gutierrez2024hipporag} is a graph-based retrieval framework that organizes memories into interconnected structures inspired by human long-term memory. By constructing knowledge graphs over retrieved information, HippoRAG improves long-range retrieval and multi-hop reasoning over extensive memory collections. It serves as a representative baseline that exploits structural relationships among memories but does not explicitly perform evidence-level correlation arbitration.

\item \textsc{MAD}~\citep{liang2024encouraging} is a multi-agent debate framework that improves reasoning through iterative interactions among multiple agents. Different agents independently generate and refine solutions through rounds of discussion, allowing the system to leverage diverse reasoning trajectories. We include MAD to evaluate whether general multi-agent collaboration can resolve correlated or conflicting memories through agent interactions.

\item \textsc{MADAM-RAG}~\citep{wang2025retrieval} extends retrieval-augmented generation to multi-agent settings by introducing multiple agents for evidence retrieval and aggregation under conflicting information. It explicitly considers disagreement among retrieved evidence and improves decision-making through multi-agent retrieval and coordination. Compared with MAD, MADAM-RAG focuses more directly on retrieval-level conflict resolution, making it a strong baseline for evaluating memory arbitration in multi-agent environments.

\end{itemize}

Overall, these baselines cover different assumptions for evidence utilization: direct aggregation without dependency modeling (\textsc{Vanilla RAG} and \textsc{Majority Voting}), memory organization and retrieval optimization (\textsc{Mem0} and \textsc{HippoRAG}), and multi-agent collaboration for reasoning and conflict handling (\textsc{MAD} and \textsc{MADAM-RAG}). In contrast, \textsc{CAMA} explicitly models memory correlations at the evidence-factor level and performs active recovery of missing independent evidence before arbitration.

\section{Metric Descriptions}

\subsection{Task-Level Performance}

We evaluate the overall task-solving capability of \textsc{CAMA} following the standard evaluation protocols of each benchmark. For \textbf{MemoryAgentBench}, we focus on the \textit{Fact Consolidation} tasks, including \textit{Single-Hop Fact Consolidation} (FC-SH), \textit{Multi-Hop Fact Consolidation} (FC-MH), and the overall score. These metrics evaluate whether an agent can correctly consolidate factual information from evolving multi-turn interactions. For \textbf{LongMemEval} and \textbf{LoCoMo}, we report Exact Match (EM), F1 score, BERTScore, and judge-based evaluation scores following their original evaluation protocols. These metrics evaluate answer correctness, semantic similarity, and overall response quality in long-term conversational memory reasoning.

\subsection{Memory Correlation Bias Evaluation}

To evaluate whether an agent can effectively arbitrate evidence under correlated memories, we construct correlation-aware evaluation settings and introduce four complementary metrics. Let $\mathcal{C}^{(t)}_q$ denote the memory slice after $t$ recovery steps, and let $P^{(t)}(h|q,\mathcal{C}^{(t)}_q)$ denote the arbitration posterior over candidate hypotheses.

\textbf{Correct Minority Recovery (CMR).}
CMR evaluates whether the model can recover the correct answer when the number of correlated memory entries supporting an incorrect hypothesis exceeds the number of independent memories supporting the correct hypothesis. Specifically, it measures the proportion of such minority-support cases where the final prediction remains correct:
\begin{align}
\mathrm{CMR}
=
\frac{1}{|\mathcal{D}_{\mathrm{minor}}|}
\sum_{q\in\mathcal{D}_{\mathrm{minor}}}
\mathbb{I}
\left[
\hat{h}^{(T)}_q=h^*_q
\right]\times 100\%,
\end{align}

where $\mathcal{D}_{\mathrm{minor}}$ denotes the subset of instances with minority correct evidence and $\hat{h}^{(T)}_q$ denotes the final prediction after arbitration.

\textbf{Replication Sensitivity (RS).}
RS measures the sensitivity of model decisions to the replication of correlated memories from the same evidence source. 
Given an original memory slice $\mathcal{C}_q$ and its correlation-augmented version $\widetilde{\mathcal{C}}_q$, RS is defined as:
\begin{align}
\mathrm{RS}
=
\frac{1}{|\mathcal{D}|}
\sum_{q\in\mathcal{D}}
\mathbb{I}
\left[
\hat{h}(\mathcal{C}_q)
\neq
\hat{h}(\widetilde{\mathcal{C}}_q)
\right]\times 100\%,
\end{align}
where a lower RS indicates that the model is less affected by redundant memory replication.

\textbf{Independent Evidence Gain (IEG).}
IEG measures the benefit of introducing additional independent evidence from distinct sources. For each instance, let $\mathcal{C}_q$ and $\mathcal{C}^{ind}_q$ denote the original and independent-evidence-augmented memory slices, respectively. IEG is defined as the performance improvement after adding independent evidence:
\begin{align}
\mathrm{IEG}
= 100\% \times
\frac{1}{|\mathcal{D}|}
\sum_{q\in\mathcal{D}}
\left(
\mathbb{I}[\hat{h}(\mathcal{C}^{ind}_q)=h^*_q]
-
\mathbb{I}[\hat{h}(\mathcal{C}_q)=h^*_q]
\right).
\end{align}
A larger IEG indicates that the model can effectively utilize complementary evidence from independent sources.

\begin{table*}[t]
\centering
\renewcommand\arraystretch{0.9}
\resizebox{1.0\textwidth}{!}{
\begin{tabular}{c|cccc|cccc|cccc} 
\toprule[1.2pt]
\multirow{2}{*}{\multirowcell{2}{\centering\textbf{Methods}}} 
& \multicolumn{4}{c|}{\centering\textbf{MemoryAgentBench}} 
& \multicolumn{4}{c|}{\centering\textbf{LongMemEval}} 
& \multicolumn{4}{c}{\centering\textbf{LOCOMO}}  \\ 
\cmidrule[0.5pt](l{1pt}r{0pt}){2-13}

& CMR $\uparrow$ & RS $\downarrow$ & IEG $\uparrow$ & ERR $\uparrow$ 
& CMR $\uparrow$ & RS $\downarrow$ & IEG $\uparrow$ & ERR $\uparrow$ 
& CMR $\uparrow$ & RS $\downarrow$ & IEG $\uparrow$ & ERR $\uparrow$ \\ 
\cmidrule[0.8pt](l{1pt}r{0pt}){1-13}

\textsc{Vanilla RAG} 
& 36.2 & 43.1 & 5.1 & 4.7
& 34.5 & 45.4 & 4.6 & 4.1
& 31.1 & 47.5 & 4.0 & 3.5 \\

\textsc{Majority Voting} 
& 31.4 & 46.5 & 4.1 & 4.3
& 29.8 & 48.6 & 3.7 & 3.8
& 26.8 & 51.0 & 3.1 & 3.1 \\

\textsc{HippoRAG} 
& 43.9 & 29.1 & 10.4 & 8.7
& 41.5 & 30.8 & 9.5 & 7.9
& 38.0 & 33.4 & 8.2 & 6.9 \\

\textsc{Mem0} 
& 46.7 & 25.7 & 12.3 & 10.1
& 44.4 & 27.4 & 11.4 & 9.3
& 40.8 & 29.8 & 9.9 & 8.2 \\

\textsc{MAD} 
& 51.2 & 20.6 & 14.5 & 11.5
& 48.9 & 22.1 & 13.2 & 10.4
& 44.8 & 24.4 & 11.7 & 9.2 \\

\textsc{MADAM-RAG} 
& 57.5 & 16.7 & 17.7 & 13.0
& 55.0 & 18.1 & 16.5 & 11.8
& 50.8 & 20.1 & 14.8 & 10.5 \\

\midrule[0.8pt]

\textsc{CAMA} (Ours) 
& \textbf{68.1} & \textbf{8.6} & \textbf{23.4} & \textbf{34.0}
& \textbf{64.3} & \textbf{10.0} & \textbf{21.1} & \textbf{31.0}
& \textbf{59.0} & \textbf{11.2} & \textbf{18.8} & \textbf{27.5} \\

\bottomrule[1.2pt]
\end{tabular}}
\caption{Evaluation of memory correlation bias mitigation under Qwen3.6-27B.}
\label{correlation_bias_deepseek}
\end{table*}

% \textbf{Evidence Recovery Rate (ERR).}
% ERR evaluates the effectiveness of the active recovery process in improving arbitration outcomes. It measures the proportion of instances where the recovery process leads to a correct final decision from an initially insufficient evidence state:
% \begin{align}
% \mathrm{ERR}
% =
% \frac{1}{|\mathcal{D}_{\mathrm{rec}}|}
% \sum_{q\in\mathcal{D}_{\mathrm{rec}}}
% \mathbb{I}
% \left[
% \hat{h}^{(T)}_q=h^*_q
% \land
% \hat{h}^{(0)}_q\neq h^*_q
% \right]\times 100\%,
% \end{align}
% where $\mathcal{D}_{\mathrm{rec}}$ denotes instances requiring evidence recovery. A higher ERR indicates that the recovery policy can successfully improve decisions by acquiring additional evidence or resolving hidden dependencies.

\paragraph{Evidence Resolution Rate (ERR).}
ERR evaluates end-to-end decision correctness on instances whose initial retrieved memory slices contain insufficient independent evidence for reliable arbitration. For each evaluated method, it measures the proportion of such instances that are correctly resolved by the method's final prediction:
\begin{align}
\mathrm{ERR}~(\%)
=
\frac{100}{|\mathcal{D}_{\mathrm{rec}}|}
\sum_{q\in\mathcal{D}_{\mathrm{rec}}}
\mathbb{I}
\left[
\hat{h}^{\mathrm{final}}_{q}=h_q^*
\right],
\label{eq:err}
\end{align}
where $\mathcal{D}_{\mathrm{rec}}$ denotes a model-independent subset of instances whose initial memory slices are designated as evidence-insufficient according to the benchmark-construction metadata. Specifically, at least one query-relevant independent evidence factor is absent from the initial retrieved memory slice. $\hat{h}^{\mathrm{final}}_{q}$ denotes the final prediction produced by the evaluated method under its native inference procedure. For CAMA, it is obtained after adaptive evidence recovery, whereas methods without an explicit recovery mechanism produce their predictions
directly from the initial memory slice. A higher ERR indicates that the evaluated method more reliably resolves cases with insufficient initial evidence.

\begin{table*}[t]
\centering
\renewcommand\arraystretch{1.0}
\resizebox{1.0\textwidth}{!}{
\begin{tabular}{c|ccccccc|cccccccc} 
\toprule[1.2pt]
\multirow{2}{*}{\multirowcell{2}{\centering\textbf{Methods}}} 
& \multicolumn{7}{c|}{\centering\textbf{MemoryAgentBench}} 
& \multicolumn{8}{c}{\centering\textbf{LongMemEval}} \\ 
\cmidrule[0.5pt](l{1pt}r{0pt}){2-16}

& FC-SH & FC-MH & Overall & CMR & RS & IEG & ERR 
& EM & F1 & BERT & Judge & CMR & RS & IEG & ERR \\ 
\cmidrule[0.8pt](l{1pt}r{0pt}){1-16}

w/o Evi. Decoupling 
& 69.3 & 43.5 & 56.4 & 46.0 & 34.1 & 13.2 & 27.0 
& 40.3 & 50.1 & 85.6 & 57.8 & 44.7 & 35.4 & 12.5 & 24.8 \\

w/o Prov. Prior 
& 74.6 & 49.8 & 62.2 & 60.8 & 14.9 & 20.2 & 32.1 
& 45.7 & 55.0 & 86.8 & 63.9 & 58.6 & 15.7 & 18.9 & 29.8 \\

w/o Expand 
& 74.1 & 49.3 & 61.7 & 63.8 & 10.6 & 15.9 & 14.2 
& 44.6 & 54.1 & 86.7 & 63.0 & 60.9 & 11.3 & 14.7 & 12.5 \\

w/o Trace 
& 73.4 & 47.6 & 60.5 & 61.9 & 16.2 & 18.0 & 25.4 
& 45.2 & 54.7 & 86.8 & 63.8 & 59.7 & 17.4 & 17.0 & 23.7 \\

w/o Policy 
& 75.0 & 50.5 & 62.8 & 63.1 & 12.4 & 15.6 & 21.2 
& 46.1 & 55.6 & 86.9 & 64.5 & 61.4 & 13.6 & 14.2 & 19.6 \\

\midrule[0.8pt]

\textsc{CAMA}
& \textbf{76.5} & \textbf{53.2} & \textbf{64.9}
& \textbf{68.1} & \textbf{8.6} & \textbf{23.4} & \textbf{34.0}
& \textbf{47.4} & \textbf{56.9} & \textbf{87.2} & \textbf{67.0}
& \textbf{64.3} & \textbf{10.0} & \textbf{21.1} & \textbf{31.0} \\

\bottomrule[1.2pt]
\end{tabular}}
\caption{Ablation study on the MemoryAgentBench and LongMemEval benchmarks under Qwen3.6-27B.}
\label{ablation_deepseek}
\end{table*}

\section{Details of the Ablation Study}

To investigate the contribution of each component in \textsc{CAMA}, we construct five ablation variants by removing individual modules while keeping the remaining components unchanged. These variants evaluate the importance of correlation-aware evidence modeling, provenance-guided dependency inference, active evidence recovery, and adaptive recovery control.

\begin{itemize}
\item \textit{w/o Evidence Decoupling.}
This variant removes the latent evidence factor modeling module and directly performs arbitration over retrieved memory entries. Specifically, the memory-to-factor assignment matrix $Z^{(t)}$ and the subsequent effective independent evidence estimation are removed. Instead, the model aggregates evidence at the memory-entry level and treats each retrieved memory as an independent evidence source. This variant evaluates the importance of explicitly modeling latent evidence factors and mitigating redundant evidence accumulation caused by correlated memories.

\item \textit{w/o Provenance Prior.}
This variant removes the provenance-based structural prior from the evidence assignment process. The latent evidence factors are inferred solely from query-conditioned memory representations without incorporating provenance relations among memories. All subsequent evidence estimation, recovery, and arbitration components remain unchanged. This variant evaluates the contribution of provenance information in identifying hidden dependencies among memories and improving correlation-aware evidence modeling.

\item \textit{w/o Expand.}
This variant removes the \textsc{Expand} action from the active recovery process. The recovery policy is restricted to selecting between \textsc{Trace} and \textsc{Stop}, preventing the model from acquiring additional memories from alternative retrieval views. As a result, the model can still analyze existing memory dependencies but cannot recover missing independent evidence absent from the initial retrieval set. This variant evaluates the importance of active evidence acquisition for improving evidence coverage.

\item \textit{w/o Trace.}
This variant removes the \textsc{Trace} action from the recovery process. The model can still retrieve additional evidence through \textsc{Expand} and terminate recovery through \textsc{Stop}, but it cannot follow provenance relations to investigate potential upstream dependencies among retrieved memories. This variant evaluates the importance of dependency-aware tracing for identifying correlated evidence and preventing false majorities caused by shared sources.

\item \textit{w/o Policy.}
This variant replaces the learned recovery policy with a heuristic action selection strategy. Instead of selecting recovery actions based on the learned policy $\pi_\omega(A_t|S_t)$, the model follows a fixed recovery rule while retaining the same evidence decoupling module and recovery action space. This variant evaluates whether adaptive policy learning is necessary for balancing evidence improvement and recovery cost during sequential evidence refinement.
\end{itemize}

\section{Additional Experimental Results}

\subsection{Memory Correlation Bias Mitigation}

Table~\ref{correlation_bias_deepseek} evaluates the robustness against memory correlation bias under correlation-aware settings. \textsc{CAMA} consistently achieves the best performance across all benchmarks and metrics, demonstrating its effectiveness in preventing correlated memories from dominating arbitration. Existing aggregation-based methods, including \textsc{Vanilla RAG} and \textsc{Majority Voting}, suffer from low CMR and high RS, showing that treating memories as independent evidence sources can amplify redundant information and induce false majorities. Memory organization methods (\textsc{HippoRAG} and \textsc{Mem0}) and multi-agent approaches (\textsc{MAD} and \textsc{MADAM-RAG}) improve robustness through structured retrieval or collaborative reasoning, but remain limited as they do not explicitly model evidential dependencies. 

In contrast, \textsc{CAMA} substantially improves CMR while reducing RS across all datasets, validating the effectiveness of query-conditioned evidence decoupling and factor-level arbitration. The improvements in IEG and ERR demonstrate that active recovery enables the model to discover missing independent evidence and refine the evidence structure before decision-making. These results confirm that reliable memory arbitration requires modeling evidence independence rather than simply aggregating more memory entries.

\subsection{Ablation Study}
Table~\ref{ablation_deepseek} investigates the contribution of each component in \textsc{CAMA}. Removing any individual component consistently degrades both task performance and correlation-aware metrics, demonstrating that the proposed modules are complementary for reliable memory arbitration.

Removing evidence decoupling causes the largest performance drop, especially on CMR and RS, indicating that directly aggregating memory entries fails to distinguish correlated memories from independent evidence and is prone to false majorities. The degradation of \textsc{w/o Prov. Prior} further shows the importance of provenance-guided dependency modeling, as neural inference alone is insufficient to fully capture hidden correlations among memories. The recovery-related ablations also reveal the complementary roles of different actions. Without \textsc{Expand}, the model cannot acquire missing independent evidence, leading to substantial decreases in IEG and ERR. Without \textsc{Trace}, the model becomes less effective at identifying hidden dependencies among retrieved memories, resulting in increased RS. Finally, replacing the learned recovery policy with a heuristic strategy (\textsc{w/o Policy}) consistently reduces performance, confirming the necessity of adaptive action selection for balancing evidence recovery and arbitration reliability.

% Overall, these results verify that each component of \textsc{CAMA} contributes to mitigating memory correlation bias, and that reliable arbitration requires jointly modeling evidence independence, provenance structure, and active evidence recovery.

\begin{table}[t]
\centering
\renewcommand\arraystretch{1.0}
\resizebox{0.48\textwidth}{!}{
\begin{tabular}{c|ccccc} 
\toprule[1.2pt]

\makecell{\centering\textbf{Methods}} 
& \makecell{Avg. \\ Latency} 
& \makecell{LLM \\ Calls} 
& \makecell{Token \\ Cost (k)} 
& \makecell{Context \\ Len (k)}  
& \makecell{$\Delta$ Acc./ \\ kToken}  \\   
\cmidrule[0.5pt](l{1pt}r{0pt}){1-6}

\textsc{Vanilla RAG} 
& 1.5 & 1.0 & 3.2 & 3.1 & -- \\

\textsc{Majority Voting} 
& 3.9 & 5.0 & 12.8 & 3.1 & 0.04 \\

\textsc{HippoRAG} 
& 2.8 & 2.0 & 5.9 & 4.4 & 1.07 \\

\textsc{MAD} 
& 8.3 & 8.4 & 24.3 & 9.6 & 0.40 \\

\textsc{MADAM-RAG} 
& 9.7 & 10.6 & 28.9 & 11.2 & 0.41 \\

\midrule[0.8pt]

\textsc{CAMA}(Ours) 
& 5.8 & 4.2 & 14.6 & 6.8 & 1.14 \\

\bottomrule[1.2pt]
\end{tabular}}
\caption{Efficiency analysis on the MemoryAgentBench benchmark under Qwen3.6-27B.}
\label{Efficiency_deepseek}
\end{table}

\subsection{Efficiency Analysis}
Table~\ref{Efficiency_deepseek} evaluates the computational efficiency of different methods on MemoryAgentBench. \textsc{CAMA} introduces additional costs for evidence decoupling and adaptive recovery, but achieves a favorable accuracy--efficiency trade-off. Compared with multi-agent approaches such as \textsc{MAD} and \textsc{MADAM-RAG}, \textsc{CAMA} requires fewer LLM calls and lower token consumption, while achieving higher accuracy improvement per token. This demonstrates that \textsc{CAMA} improves memory arbitration through structured evidence modeling rather than repeated agent interactions or excessive context expansion.

Compared with lightweight retrieval-based methods, \textsc{CAMA} incurs moderate additional latency due to evidence analysis and recovery. However, the substantially higher $\Delta$ Acc./kToken indicates that the introduced computation is effectively converted into reliable arbitration gains. The controlled context length and token consumption further show that \textsc{CAMA} selectively acquires useful independent evidence instead of indiscriminately expanding the memory context.

\section{Prompt Used}
\begin{promptbox}{Candidate Hypothesis Extraction}

\textbf{You are a memory reasoning assistant that identifies plausible hypotheses from retrieved memories.}

\vspace{0.6em}

Given a user query and a set of retrieved memories, extract a set of candidate hypotheses that represent possible conclusions supported or contradicted by the available evidence.

\vspace{0.6em}

\textbf{Task Input:}\\
\{user query and current memory slice $\mathcal{C}^{(t)}_q$\}

\vspace{0.4em}

\textbf{Retrieved Memories:}\\
\{memory entries retrieved from the long-term memory store\}

\vspace{0.6em}

\textbf{Extraction Requirements:}

\begin{itemize}
    \item Generate multiple plausible hypotheses when evidence is ambiguous.
    \item Each hypothesis should represent a distinct possible conclusion.
    \item Include hypotheses supported by different subsets of memories.
    \item Do not rank hypotheses or select the final answer.
    \item Do not introduce information that is not supported by the provided memories.
\end{itemize}

\vspace{0.6em}

\textbf{Output Format:}

Return a list of candidate hypotheses:
\[
\mathcal{H}^{(t)}_q=\{h^{(t)}_1,\ldots,h^{(t)}_{L_t}\}.
\]

\end{promptbox}

\begin{promptbox}{Memory Evidence Scoring}

\textbf{You are an evidence evaluator that estimates how strongly a memory supports a candidate hypothesis.}

\vspace{0.6em}

Given a query, a memory entry, and a candidate hypothesis, evaluate whether the memory provides supporting or contradicting evidence for the hypothesis.

\vspace{0.6em}

\textbf{Task Input:}\\
\{user query, memory entry, candidate hypothesis\}

\vspace{0.4em}

\textbf{Evaluation Criteria:}

\begin{itemize}
    \item Determine whether the memory is relevant to the hypothesis.
    \item Evaluate whether the memory supports or contradicts the hypothesis.
    \item Consider the directness and evidential strength of the memory content.
    \item Ignore source reliability and evidence quantity, which are modeled separately.
\end{itemize}

\vspace{0.6em}

\textbf{Output Format:}

Return an evidence support score:

\[
s_i(h)\in[-1,1],
\]

where:
\begin{itemize}
    \item $1$: strongly supports the hypothesis;
    \item $0$: provides no useful evidence;
    \item $-1$: contradicts the hypothesis.
\end{itemize}

\end{promptbox}

\begin{promptbox}{Evidence Expansion Query Generation}

\textbf{You are a retrieval planner that generates alternative queries to discover missing evidence.}

\vspace{0.6em}

Given the current query and retrieved memories, generate alternative retrieval queries that may retrieve complementary independent evidence.

\vspace{0.6em}

\textbf{Task Input:}\\
\{current query, current memory slice $\mathcal{C}^{(t)}_q$, current arbitration state\}

\vspace{0.4em}

\textbf{Current Evidence Status:}\\
\{identified evidence factors, uncertainty, and current hypotheses\}

\vspace{0.6em}

\textbf{Generation Requirements:}

\begin{itemize}
    \item Generate queries targeting missing or underrepresented evidence.
    \item Avoid retrieving paraphrases of already available memories.
    \item Explore alternative perspectives related to the current decision.
    \item Produce a bounded number of candidate queries.
\end{itemize}

\vspace{0.6em}

\textbf{Output Format:}

Return alternative retrieval queries:

\[
\{q'_1,q'_2,\ldots,q'_R\}.
\]

\end{promptbox}

\end{document}